\documentclass[sn-mathphys-num]{sn-jnl}

\usepackage{graphicx}%
\usepackage{multirow}%
\usepackage{amsmath,amssymb,amsfonts}%
\usepackage{amsthm}%
\usepackage{mathrsfs}
\usepackage[title]{appendix}%
\usepackage{xcolor}%
\usepackage{textcomp}%
\usepackage{manyfoot}%
\usepackage{booktabs}%
\usepackage{algorithm}%
\usepackage{algorithmicx}%
\usepackage{algpseudocode}%
\usepackage{listings}%
\usepackage{multicol}
\usepackage{calc}
\usepackage{tabularx}
\usepackage{CJKutf8}
\usepackage{makecell}
\usepackage{fdsymbol}
\usepackage{color}
\usepackage{colortbl} 

\usepackage{adjustbox}
\definecolor{aliceblue}{RGB}{178, 217, 245}
\newcommand{\CC}{\cellcolor{aliceblue}}
\definecolor{babyblue}{RGB}{217, 239, 251}
\newcommand{\CB}{\cellcolor{babyblue}}
\definecolor{lightblue}{RGB}{157,195,230}
\definecolor{pinkred}{RGB}{255,179,167}

\begin{document}

\title[Article Title]{LLMs for Knowledge Graph Construction and  Reasoning: Recent Capabilities and Future Opportunities}


\author[1,2]{\fnm{Yuqi} \sur{Zhu}}\email{zhuyuqi@zju.edu.cn}
\equalcont{These authors contributed equally to this work.}

\author[1,2]{\fnm{Xiaohan} \sur{Wang}}
\equalcont{These authors contributed equally to this work.}

\author[1,2]{\fnm{Jing} \sur{Chen}}
\equalcont{These authors contributed equally to this work.}

\author[1,2]{\fnm{Shuofei} \sur{Qiao}}
\author[1,2]{\fnm{Yixin} \sur{Ou}}
\author[1,2]{\fnm{Yunzhi} \sur{Yao}}
\author[3]{\fnm{Shumin} \sur{Deng}}
\author[1,2]{\fnm{Huajun} \sur{Chen}}
\author*[1,2]{\fnm{Ningyu} \sur{Zhang}}\email{zhangningyu@zju.edu.cn}




\affil*[1]{\orgname{Zhejiang University}, \country{China}}

\affil[2]{\orgname{ZJU-Ant Group Joint Research Center for Knowledge Graphs}, \country{China}}                                                 

\affil[3]{\orgname{National University of Singapore, NUS-NCS Joint Lab}, \country{Singapore}}


\abstract{This paper presents an exhaustive quantitative and qualitative evaluation of Large Language Models (LLMs) for Knowledge Graph (KG) construction and reasoning. We engage in experiments across eight diverse datasets, focusing on four representative tasks encompassing entity and relation extraction, event extraction, link prediction, and question-answering, thereby thoroughly exploring LLMs’ performance in the domain of construction and inference. Empirically, our findings suggest that LLMs, represented by GPT-4, are more suited as inference assistants rather than few-shot information extractors. Specifically, while GPT-4 exhibits good performance in tasks related to KG construction, it excels further in reasoning tasks, surpassing fine-tuned models in certain cases. Moreover, our investigation extends to the potential generalization ability of LLMs for information extraction, leading to the proposition of a Virtual Knowledge Extraction task and the development of the corresponding VINE dataset. Based on these empirical findings, we further propose \textbf{AutoKG}, a multi-agent-based approach employing LLMs and external sources for KG construction and reasoning. We anticipate that this research can provide invaluable insights for future undertakings in the field of knowledge graphs.}

\keywords{Knowledge Graph, Information Extraction, GPT-4, Large Language Model}



\maketitle
\section{Introduction}\label{sec1}
Knowledge Graph (KG) is a semantic network comprising entities, concepts, and relations  \cite{cai2022temporal,zhu2022multi,DBLP:journals/corr/abs-2212-05767,DBLP:journals/corr/abs-2305-08698,Unify_LLM_and_KG,LLM_and_KG_Opportunities_and_KG}, which can catalyse applications across various scenarios. 
Constructing KGs~\cite{DBLP:conf/emnlp/Ye0CC22,kgc/Theme-specific} typically involves multiple tasks such as 
Named Entity Recognition (NER)~\cite{DBLP:journals/tacl/ChiuN16,ie/IEPile}, Relation Extraction (RE)~\cite{DBLP:conf/emnlp/ZengLC015,DBLP:conf/www/ChenZXDYTHSC22}, Event Extraction (EE)~\cite{DBLP:conf/acl/ChenXLZ015,DBLP:conf/wsdm/DengZKZZC20}, and Entity Linking (EL)~\cite{DBLP:journals/tkde/ShenWH15}.
Additionally, Link Prediction (LP)~\cite{DBLP:conf/aaai/ZhangDKSS18, DBLP:journals/tkdd/RossiBFMM21} is a crucial step for KG reasoning, essential for understanding constructed KGs. 
These KGs also hold a central position in Question Answering (QA) tasks~\cite{DBLP:conf/emnlp/KarpukhinOMLWEC20,DBLP:journals/corr/abs-2101-00774}, especially in conducting inference based on question context, involving the construction and application of relation subgraphs. 
This paper empirically investigates the potential applicability of LLMs in the KG domain, taking ChatGPT and GPT-4~\cite{DBLP:journals/corr/abs-2303-08774} as examples. 
The research begins with an examination of the fundamental capabilities of LLMs \cite{DBLP:journals/corr/abs-2303-13547,DBLP:journals/corr/abs-2302-13814,DBLP:journals/corr/abs-2304-05613,DBLP:journals/corr/abs-2303-18223}, progressing to explore possible future developments, aiming to enhance our understanding of LLMs and introduce new perspectives and methods to the field of knowledge graphs.

\begin{figure*}[!ht]

\centering 
\includegraphics[width=1\textwidth]{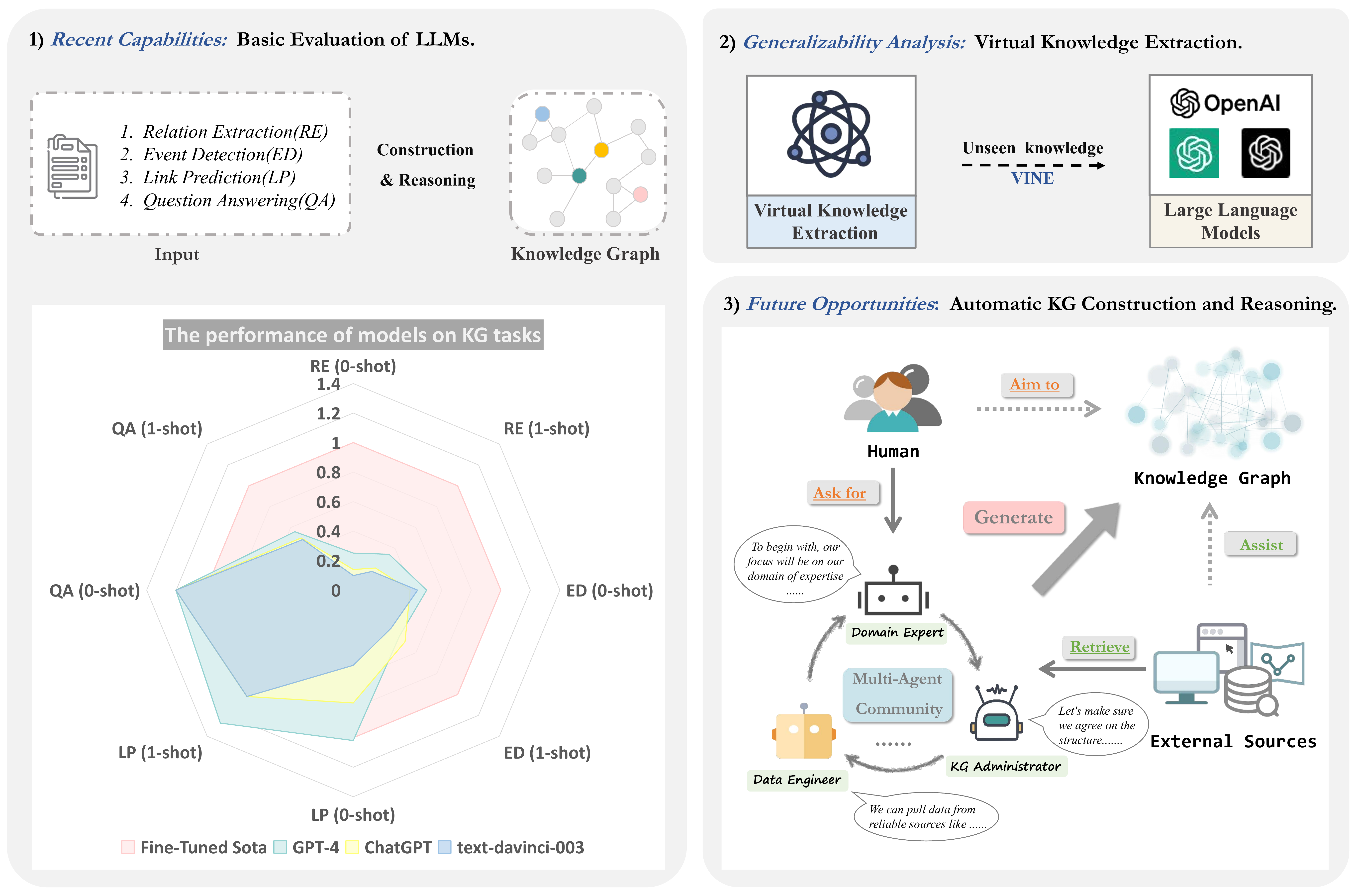} 
\caption{The overview of our work. There are three main components: 1) \textit{\textbf{Basic Evaluation}}: detailing our assessment of large models (text-davinci-003, ChatGPT, and GPT-4), in both zero-shot and one-shot settings, using performance from fully supervised state-of-the-art models as benchmarks; 2) \textit{\textbf{Virtual Knowledge Extraction}}: an examination of LLMs' virtual knowledge capabilities on the constructed VINE dataset; and 3) \textit{\textbf{Automatic KG}}: the proposal of utilizing multiple agents to facilitate the construction and reasoning of KGs.}
\label{fig:overall}
\end{figure*}

\textit{\textbf{Recent Capabilities.}} 
Entity and Relation Extraction, along with Event Extraction, are pivotal for Knowledge Graph (KG) construction tasks~\cite{DBLP:journals/corr/abs-2302-10205,eval/Capabilities,DBLP:conf/emnlp/LiWK23,DBLP:conf/emnlp/WanCMLSLK23}. 
They play a critical role in organizing vast amounts of entity, relation, and event data into structured representations, forming the foundational elements that underpin the construction and enrichment of KGs.
Meanwhile, Link Prediction, as a core task of KG reasoning~\cite{reason/nlptasksolver}, aims to uncover latent relationships between entities, thereby enriching the knowledge graph.
Additionally, we delve into the utilization of LLMs in knowledge-based Question Answering tasks~\cite{reason/logicalreasoning, kgr/kgagent} to gain a thorough insight into their reasoning capabilities.
Given these considerations, we select these tasks as representatives for evaluating both the construction and reasoning of KGs. 
As illustrated in Figure~\ref{fig:overall}, 
our initial investigation targets the zero-shot and one-shot abilities of large language models across the aforementioned tasks.
This analysis serves to assess the potential usage of such models in the field of knowledge graphs.
The empirical findings reveal that LLMs like GPT-4 exhibit limited effectiveness as \textit{a few-shot information extractor}, yet demonstrate considerable proficiency as \textit{an inference assistant}.

\textit{\textbf{Generalizability Analysis.}} To delve deeper into the behavior of LLMs in information extraction tasks, we devise a unique task termed \textbf{Virtual Knowledge Extraction}, targeting LLMs' ability to generalize and extract unfamiliar knowledge. 
This undertaking aims to discern whether the observed performance enhancements on these tasks are attributed to the extensive internal knowledge repositories of LLMs or to their potent generalization capabilities facilitated by instruction tuning~\cite{thenflancollection} and Reinforcement Learning from Human Feedback (RLHF) \cite{DBLP:conf/nips/ChristianoLBMLA17}. 
And our experiments on a newly constructed dataset, \textbf{VINE}, indicate that large language models like GPT-4 can acquire new knowledge from instructions and effectively execute extraction tasks, thereby affording a more nuanced understanding of large models to a certain extent.

\textit{\textbf{Future Opportunities.}}
In light of the preceding experiments, we further examine prospective directions for knowledge graphs.
Given the remarkable generalization capabilities of large models~\cite{chatgpt_analysis,survey/harness_power_of_llm}, we opt to employ them to aid in the construction of KG. 
Compared to smaller models, these LLMs mitigate potential resource wastage and demonstrate notable adaptability in novel or data-scarce situations. 
However, it's important to recognize their strong dependence on prompt engineering~\cite{cot} and the inherent limitations of their knowledge cutoff.
Consequently, researchers are exploring interactive mechanisms that allow LLMs to access and leverage external resources, aiming to enhance their performance further~\cite{CoRR23_INLP}.

On this basis, we introduce the concept of AutoKG - autonomous KG construction and reasoning via multi-agent communication.
In this framework, the human role is diminished, with multiple communicative agents each playing their respective roles. 
These agents interact with external sources, collaboratively accomplishing the task.
We summarize our contributions as follows~\footnote{The code and datasets are in \url{https://github.com/zjunlp/AutoKG}.}:
 \begin{itemize}
   \item  We evaluate LLMs, including ChatGPT and GPT-4, offering an initial understanding of their capabilities by evaluating their zero-shot and one-shot performance on KG construction and reasoning on eight benchmark datasets.
   \item  We design a novel Virtual Knowledge Extraction task and construct the \textbf{VINE} dataset.
   By evaluating the performance of LLMs on it, we further demonstrate that LLMs such as GPT-4 possess strong generalization abilities.
   \item We introduce the concept of 
   automatic KG construction and reasoning, known as \textbf{AutoKG}. 
   Leveraging LLMs' inner knowledge, we enable multiple agents of LLMs to assist in the process through iterative dialogues, providing insights for future research.
 \end{itemize}

\section{Recent Capabilities of LLMs for KG Construction and Reasoning}
The release of large language models like GPT-4, recognized for their remarkable general capabilities, has been considered by researchers as the spark of artificial general intelligence (AGI)~\cite{bubeck2023sparks}.
To facilitate an in-depth understanding of their performance in KG-related tasks, a series of evaluations are conducted. 
\S\ref{sec:evaluation_principle} introduces the evaluation principles, followed by a detailed analysis in \S\ref{sec:kg_construction_reasoning} on the performance of LLMs in the construction and reasoning tasks, highlighting variations across different datasets and domains.
Moreover,~\S\ref{sec:why_perfrom_well} delves into the reasons underlying the subpar performance of LLMs in certain tasks. 
And finally, \S\ref{sec:virtual_knowledge_extraction} discusses whether the models’ performance is genuinely indicative of generalization abilities or influenced by inherent advantages of the knowledge base.

\subsection{Evaluation Principle}
\label{sec:evaluation_principle}
In this study, we conduct a comprehensive assessment of LLMs, represented by GPT-4, and specifically analyze the performance disparities and enhancements between GPT-4 and other models in the GPT series, such as ChatGPT. 
A primary area of investigation is the models' performance in zero-shot and one-shot tasks, as these tasks illuminate the models' generalization capabilities under data-limited conditions. 
Given that some experiments in our study rely on randomly sampled subsets of datasets, it is important to note that there may be inherent variability in the results due to this sampling approach.
We deliberately choose zero-shot and one-shot tasks over those requiring more examples, as they better test the models' adaptability and practical application in scenarios with sparse data. 
The experimental prompt used is detailed in Appendix \ref{sec:Prompts}. 
Utilizing the evaluation results, our objective is to explore the reasons behind the models' exemplary performance in specific tasks and identify potential areas of improvement. 
Ultimately, our goal is to derive valuable insights for future advancements in such models.

\subsection{KG Construction and Reasoning}
\label{sec:kg_construction_reasoning}
\subsubsection{Settings}
\textit{\textbf{Datasets.}}
During the task of Entity and Relation Extraction, Event Extraction, we employ DuIE2.0~\cite{DBLP:conf/nlpcc/LiHSJLJZLZ19}, SciERC~\cite{DBLP:conf/emnlp/LuanHOH18}, Re-TACRED~\cite{DBLP:conf/aaai/StoicaPP21}, and MAVEN~\cite{DBLP:conf/emnlp/WangWHJHLLLLZ20} datasets. 
For Link Prediction, we utilized FB15K-237~\cite{DBLP:conf/emnlp/ToutanovaCPPCG15} and ATOMIC 2020~\cite{DBLP:conf/aaai/HwangBBDSBC21} datasets. 
Finally, FreebaseQA~\cite{DBLP:conf/naacl/JiangW019} and MetaQA~\cite{DBLP:conf/aaai/ZhangDKSS18} datasets are used in the Question Answering task. 
The dataset used is described in detail in Appendix \ref{sec:datasets}.

\begin{figure*}[!ht]

\centering 
\includegraphics[width=1\textwidth]{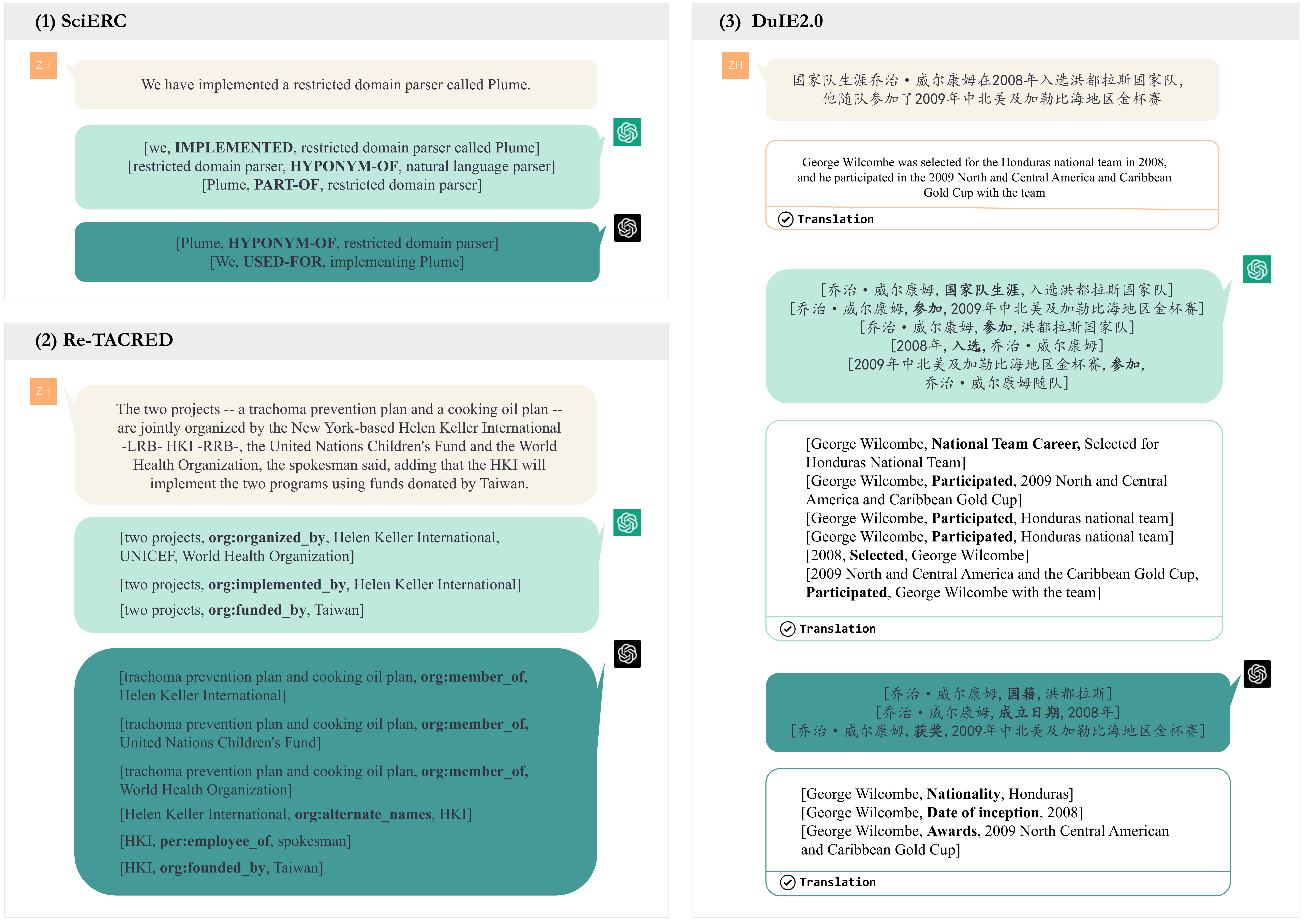} 
\caption{Examples of ChatGPT and GPT-4 on the RE datasets.~(1)~Zero-shot on the SciERC dataset~(2)~Zero-shot on the Re-TACRED dataset~(3)~One-shot on the DuIE2.0 dataset}
\label{fig:dialogue}
\end{figure*}

\subsubsection{Overall Results}

\textit{\textbf{Entity and Relation Extraction.}}
We conduct experiments on DuIE2.0, Re-TACRED, and SciERC, each involving 20 samples in the test/valid sets, encompassing all types of relationships present within the datasets. 
Here we use PaddleNLP LIC2021 IE
~\footnote{\url{https://github.com/PaddlePaddle/PaddleNLP/tree/develop/examples/information_extraction/DuIE}.},
PL-Marker \cite{DBLP:conf/acl/YeL0S22} and EXOBRAIN~\cite{DBLP:journals/corr/abs-2107-09332} as baselines on each dataset, respectively.
Concurrently, for evaluation purposes, the results are reported utilizing the standard micro F1 score.
As shown in Table~\ref{tab:KG-construct-reason}, GPT-4 performs relatively well in both zero-shot and one-shot manners compared to ChatGPT, even though its performance has not yet surpassed that of fully supervised small models.

$\bullet$~\textbf{Zero-shot}
GPT-4's zero-shot performance significantly improves across all tested datasets, especially in DuIE2.0, scoring 31.03, compared to ChatGPT's 10.3.
Specifically, in the example of Re-TACRED in Figure~\ref{fig:dialogue}, ChatGPT fails to extract the target triple, possibly due to the close proximity of head and tail entities and the ambiguity of predicates. 
In contrast, GPT-4 gives the correct answer `` org:alternate$\_$names'', highlighting its superior language comprehension.

$\bullet$~\textbf{One-shot}
Simultaneously, the optimization of text instructions has been shown to enhance the performance of LLMs.
In the context of DuIE2.0 shown in Figure~\ref{fig:dialogue}, 
GPT-4 discerns an implicit relation from a statement about George Wilcombe's association with the Honduras national team. 
This precision is attributed to GPT-4's extensive knowledge base, which facilitates the inference of George Wilcombe's nationality.
However, it is also observed that GPT-4 encounters challenges with complex sentences, with factors such as prompt quality and relational ambiguity affecting the outcomes.

\begin{figure*}[!ht]
\centering 
\includegraphics[width=\textwidth]{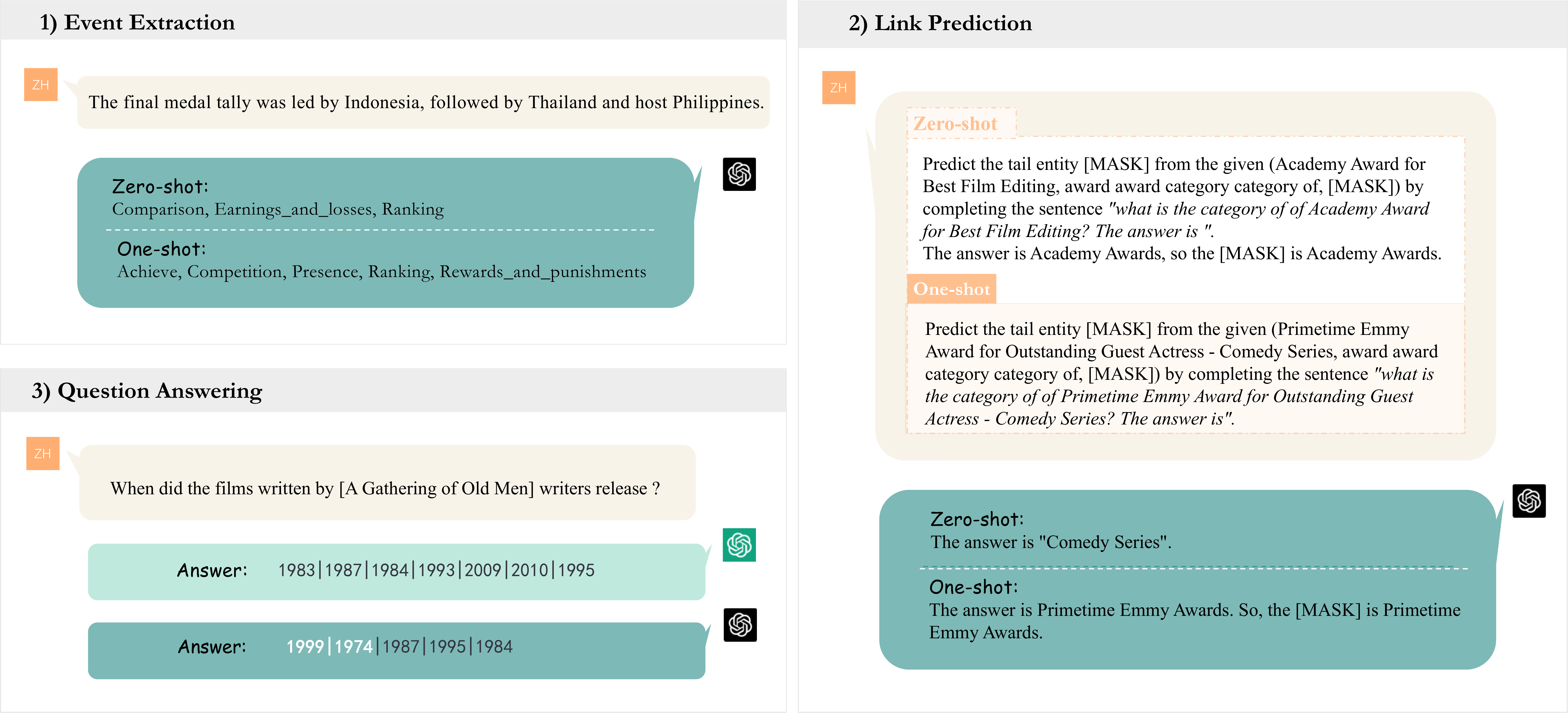} 
\caption{Here are examples of task Event Extraction, Link Prediction and Question Answering.}
\label{fig:KG_resoning_example}
\end{figure*}

\textit{\textbf{Event Extraction.}}
For simplification, we conduct event detection experiments on 20 random samples from MAVEN, encompassing all event types.
Using the F-score metric, GPT-4's performance is benchmarked against the existing state-of-the-art (SOTA)~\cite{DBLP:journals/corr/abs-2204-07241} model, as well as to other models in the GPT family. 
Based on our results, GPT-4 shows inconsistent superiority over the SOTA, with both GPT-4 and ChatGPT outperforming each other in different scenarios.

$\bullet$~\textbf{Zero-shot}
As shown in Table~\ref{tab:KG-construct-reason}, GPT-4 outperforms ChatGPT. 
For the sentence ``Now an established member of the line-up, he agreed to sing it more often.'', ChatGPT generates the result \textit{Becoming$\_$a$\_$member}, while GPT-4 identifies two more: \textit{Agree$\_$or$\_$refuse$\_$to$\_$act, Performing}.
It is worth noting that in this experiment, ChatGPT frequently provides answers with a single event type. 
In contrast, GPT-4's ability to grasp complex contextual information enables it to identify multiple event types within these sentences.

$\bullet$~\textbf{One-shot}
In this configuration, ChatGPT's performance improves notably, while GPT-4 experiences a slight decline.
Figure~\ref{fig:KG_resoning_example} illustrates that GPT-4 incorrectly identifies five event types where the correct answers are \textit{Process$\_$end} and \textit{Come$\_$together}.
Despite detecting underlying ranking and comparison information, GPT-4 misses the trigger words \textit{final} and \textit{host}.
Simultaneously, we observe that under one-shot setup, GPT-4 tends to produce a higher number of erroneous responses when it is unable to identify the correct ones. 
We theorize this could stem from implicit type indications of the dataset.

\begin{table*}[!htbp]
 \centering
 \small
 \caption{KG Construction and KG Reasoning tasks.}
  \scalebox{0.7}{
 \begin{tabular}{l|cccc|cccc}
 \toprule
    \multirow{3}{*}{ \textbf{Model}}   
    & \multicolumn{4}{c}{\multirow{2}{*}{\textbf{Knowledge Graph Construction}}}
    & \multicolumn{4}{c}{\multirow{2}{*}{\textbf{Knowledge Graph Reasoning}}}
    \\\\
    \cmidrule(lr){2-5}  \cmidrule(lr){6-9}
             & DuIE2.0  & Re-TACRED     &SciERC   & MAVEN    & FB15K-237 & ATOMIC2020  &FreebaseQA &MetaQA \\\midrule
     Fine-Tuned SOTA     &\CC69.42  &\CC91.4  &\CC53.2  &\CC68.8  &\CB32.4  &\CC46.9  &\CB79.0  &\CC100\\\midrule
    \multicolumn{9}{c}{\textbf{Zero-shot}}\\\midrule
    text-davinci-003  &11.43  &9.8  &4.0  &30.0   &16.0  &15.1  &\CC95.0  &33.9     \\
    ChatGPT  &10.26  &15.2   &4.4  &26.5     &24.0  &10.6   &\CC95.0  &52.7      \\
    GPT-4  &31.03   &15.5   &7.2   &\CB34.2  &32.0   &16.3   &\CC95.0   &\CB63.8       \\\midrule
     \multicolumn{9}{c}{\textbf{One-shot}}\\\midrule
    text-davinci-003 &30.63  &12.8  &4.8  &25.0   &32.0  &14.1  &\CC95.0  &49.5      \\
    ChatGPT &25.86   &14.2  &5.3  &34.1          &32.0   &11.1   &\CC95.0   &50.0       \\
    GPT-4 & \CB41.91  &\CB22.5    &\CB9.1 &30.4    &\CC40.0   &\CB19.1   &\CC95.0   &56.0     \\
    \bottomrule
 \end{tabular}
 }
 \label{tab:KG-construct-reason}
\end{table*}

\textit{\textbf{Link Prediction.}}
Task link prediction involves experiments on two distinct datasets, FB15k-237 and ATOMIC2020. 
The former is a random sample set comprising 25 instances, whereas the latter encompasses 23 instances on behalf of all possible relations. 
Among various approaches, the best performing fine-tuned models are C-LMKE (BERT-base)~\cite{wang2022language} and COMET (BART)~\cite{Hwang2021COMETATOMIC2O} for each.

$\bullet$~\textbf{Zero-shot}
In Table~\ref{tab:KG-construct-reason}, GPT-4 on the FB15k-237 demonstrates that its hits@1 score is nearing the SOTA level. 
Regarding the ATOMIC2020, while GPT-4 still exceeds the other two models, there remains a considerable discrepancy in terms of bleu1 score between GPT-4's performance and the fine-tuned SOTA achieved.
In the zero-shot context, it is observable that ChatGPT often refrains from providing immediate answers when faced with link prediction ambiguity, opting instead to seek further contextual data.  
This cautious approach contrasts with GPT-4's propensity to offer direct responses, suggesting possible differences in their reasoning and decision-making strategies.

$\bullet$~\textbf{One-shot}
Instructional text optimization has proven beneficial in enhancing GPT series' performance in link prediction tasks. 
Empirical evaluations demonstrate one-shot GPT-4 improves results on both datasets, supporting accurate tail entity prediction in triples. 
In the example of Figure~\ref{fig:KG_resoning_example}, the target \textit{[MASK]} is \textit{Primetime Emmy Award}. 
In zero-shot setting, GPT-4 fails to comprehend the relation, leading to an incorrect response \textit{Comedy Series}. 
However, when the demonstration is incorporated, GPT-4 successfully identifies the target.

\textit{\textbf{Question Answering.}}
We conduct the evaluation using two prevalent Knowledge Base Question Answering datasets, FreebaseQA and MetaQA, with 20 random instances sampled from each. 
In MetaQA, we sample proportional to their dataset representation. 
\citet{DBLP:journals/corr/abs-2210-00063} and ~\citet{DBLP:journals/corr/abs-2303-02206} represent the SOTA models employed. 
And for both datasets, AnswerExactMatch is adopted as the metric of evaluation.

$\bullet$~\textbf{Zero-shot}
As shown in Table~\ref{tab:KG-construct-reason}, ChatGPT and GPT-4 demonstrate identical performance on FreebaseQA, surpassing preceding fully supervised SOTA by 16$\%$.
Yet, no advantage of GPT-4 over ChatGPT is observed. 
For MetaQA, there is still a large gap between LLMs and supervised SOTA, possibly due to multi-answer questions and LLM input token constraints. 
Nevertheless, GPT-4 outperforms ChatGPT by 11.1 points, which indicates the superiority of GPT-4 against ChatGPT on more challenging QA tasks. 
Specifically, in the example of Figure~\ref{fig:KG_resoning_example}, GPT-4 correctly answers a multi-hop question from MetaQA, yielding both \textit{1999} and \textit{1974} release dates, highlighting its superior performance in multi-hop QA tasks over ChatGPT.

$\bullet$~\textbf{One-shot}
We also conduct experiments under one-shot setting by randomly sampling one example from the train set as the in-context exemplar. 
Results in Table~\ref{tab:KG-construct-reason} demonstrate that only text-davinci-003 benefits from the prompt, while both ChatGPT and GPT-4 encounter a performance drop. 
This can be attributed to the notorious alignment tax where models sacrifice some of their in-context learning ability for aligning with human feedback.

\subsubsection{KG Construction vs. Reasoning}
\label{sec:kg_constructandreason}
Our experiments on KG construction and reasoning reveal that LLMs exhibit superior reasoning skills compared to their construction capabilities. 
Given the challenge of quantifying reasoning and construction abilities, we assess the comparative capabilities of LLMs in these tasks by measuring the performance differential between LLMs and the current SOTA methodologies. 
Larger performance disparities indicate poorer performance.
Despite the exemplary performance of LLMs, they do not surpass the current state-of-the-art models in KG construction under zero-shot and one-shot settings, indicating limitations in extracting information from sparse data.
Conversely, all LLMs in one-shot, and GPT-4 in zero-shot, match or near SOTA performance on the FreebaseQA and FB15K-237 datasets.
Moreover, they exhibit relatively good performance across the remaining datasets, which underscores their adaptability in KG reasoning tasks as well.
The intrinsic complexity of KG construction tasks may account for this discrepancy in performance. 
Furthermore, the robust reasoning performance of LLMs might be attributed to their exposure to relevant knowledge during pre-training.

\subsubsection{General vs. Specific Domain}

In our study, we evaluate the performance of large language models, exemplified by GPT-4, across diverse knowledge domains, ensuring a balanced assessment in both generic and specialized contexts. 
We employed a consistent method of evaluating relative task capabilities, similar to the performance disparity assessment described in \S\ref{sec:kg_constructandreason}. 
The chosen benchmarks, SciERC and Re-TACRED, represent scientific and general domains, respectively. 
While Re-TACRED exhibits a broader range of relation types compared to the seven in SciERC, both GPT-4 and ChatGPT underperform on the specialized SciERC dataset, indicating their limitations in domain-specific data. 
Interestingly, GPT-4's performance boost on SciERC is less pronounced than on Re-TACRED when given one demonstration.
Specifically, during our experiments, we note challenges in LLMs' recognition and understanding of specialized terms within the SciERC dataset.
We hypothesize that the subpar performance on specialized datasets may stem from these models being predominantly trained on vast general corpora, thereby lacking sufficient domain-specific expertise.

\subsection{Discussion: Why LLMs do not present satisfactory performance on some tasks?}
\label{sec:why_perfrom_well}
Our experiments underscore GPT-4's ability to extract knowledge across diverse domains, albeit not surpassing the performance of fine-tuned models.
This observation also aligns with findings from previous research ~\cite{DBLP:journals/corr/abs-2302-10205,DBLP:journals/corr/abs-2303-03836}. 
Our experiment, conducted in March-April 2023, uses an interactive interface rather than an API to evaluate the GPT models on a randomly selected subset of datasets.

Notably, in assessing large models across eight datasets, we identify that the outcomes may be subject to various factors.
\textbf{\textit{Dataset Quality}}: 
Using the KG construction task as an illustration, dataset noise could lead to ambiguities.
Complex contexts and potential label inaccuracies may also negatively impact model evaluation.
\textbf{\textit{Instruction Quality}}: 
Model performance is notably influenced by the semantic depth of instructions.
Finding optimal instructions through prompt engineering~\footnote{\url{https://www.kdnuggets.com/publications/sheets/ChatGPT_Cheatsheet_Costa.pdf}} can enhance performance. 
An In-context Learning~\cite{DBLP:journals/corr/abs-2301-00234} approach with relevant samples can further improve outcomes.
\textbf{\textit{Evaluation Methods}}: 
Current methods may not be entirely apt for assessing the capabilities of large models like ChatGPT and GPT-4.
Dataset labels may not capture all correct responses, and answers involving synonymous terms might not be accurately recognized.

\subsection{Discussion: Do LLMs have memorized knowledge or truly have the generalization ability?}
\label{sec:virtual_knowledge_extraction}
Leveraging insights from prior studies, it is apparent that large models are adept at swiftly extracting structured knowledge from minimal information.
This observation raises a question regarding the origin of the performance advantage in LLMs: is it due to the substantial volume of textual data used in pre-training phases, enabling the models to acquire pertinent knowledge, or is it attributed to their robust inference and generalization capabilities?
\begin{figure}[!ht]
\centering 
\includegraphics[width=0.60\textwidth]{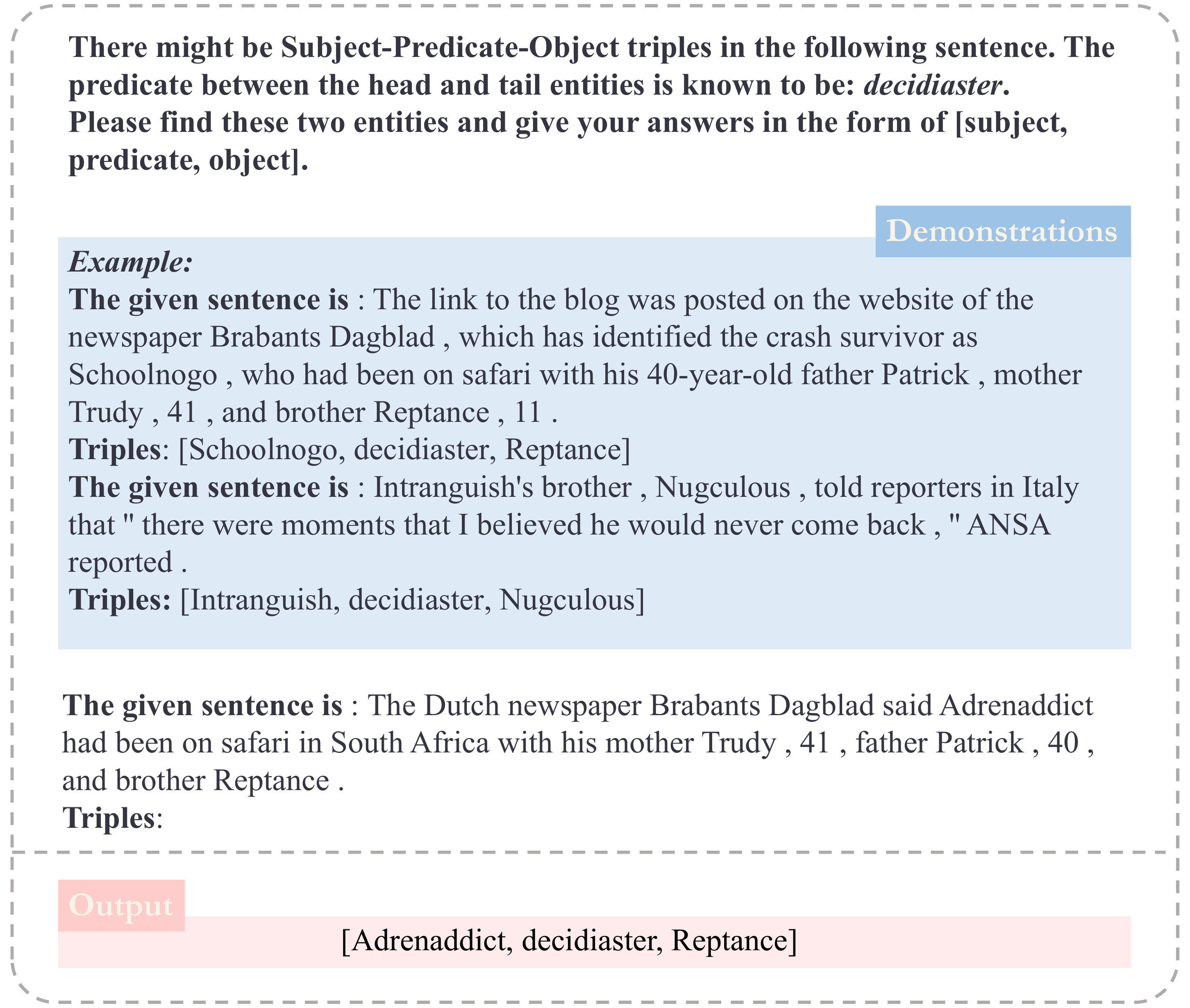} 
\caption{Prompts used in Virtual Knowledge Extraction.
\textcolor{lightblue}{The blue box} is the demonstration and \textcolor{pinkred}{the pink box} is the corresponding answer.}
\label{fig:virtual_extract}
\end{figure}

To explore this, we design the \textbf{Virtual Knowledge Extraction} task, targeting LLMs' ability to generalize and extract unfamiliar knowledge. 
Unlike conventional benchmarks, the task focuses on evaluating how models perform when confronted with information they have not previously encountered, rather than relying solely on knowledge accumulated during pre-training. 
Existing datasets largely comprise entities familiar to LLMs, potentially sourced from their pre-training corpora, thereby possibly including relationships already encoded within these corpora during extraction tasks.
Addressing these dataset constraints, we introduce \textbf{VINE}, a novel dataset specifically crafted for Virtual Knowledge Extraction.

In VINE, we fabricate entities and relations not found in reality, structuring them into knowledge triples.
We then instruct the models to extract this synthetic knowledge, 
using the efficiency of this process as an indicator of LLMs' capacity to manage virtual knowledge. 
It is worth noting that we construct VINE based on the test set of Re-TACRED. 
The primary idea behind this process is to replace existing entities and relations in the original dataset with unseen ones, thereby creating unique virtual knowledge scenarios.

\subsubsection{Data Collection}
Considering the vast training datasets of large models like GPT-4, it is challenging for us to find the knowledge that they do not recognize.
Using GPT-4 data up to September 2021 as a basis, we select a portion of participants' responses from two competitions organized by \textit{the New York Times} in 2022
and 2023
as part of our data sources.

However, due to the limited number of responses in the above contests and to enhance data source diversity, we also create new words by randomly generating letter sequences.
This is accomplished by generating random sequences between 7 to 9 characters in length (including 26 letters of the alphabet and the symbol ``\_'') and appending common noun suffixes at random to finalize the construction. 
More details could be found in Appendix~\ref{sec:data_collect}.

\subsubsection{Preliminary Results}
In our experiment, we conduct a random selection of ten sentences for evaluation, ensuring they encompass all relationships.
We assess the performance of ChatGPT and GPT-4 on these test samples after learning two demonstrations of the same relation. 
Notably, GPT-4 successfully extracted 80$\%$ of the virtual triples, while the accuracy of ChatGPT is only 27$\%$.

In Figure~\ref{fig:virtual_extract}, we provide large models with a triple composed of virtual relation types and virtual head and tail entities—\textit{[Schoolnogo, decidiaster, Reptance]} and~\textit{[Intranguish, decidiaster, Nugculous]}—along with the respective demonstrations. 
The results demonstrate that GPT-4 effectively completed the extraction of the virtual triple.
Consequently, we tentatively conclude that GPT-4 exhibits a relatively strong generalization ability and can rapidly acquire the capability to extract new knowledge through instructions, rather than relying solely on the memory of relevant knowledge. 
Related work~\cite{DBLP:journals/corr/abs-2303-03846} has also confirmed that large models possess an exceptionally strong generalization ability concerning instructions.

\section{Future Opportunities: Automatic KG Construction and Reasoning}
In contemplating the trajectory of Knowledge Graph, the pronounced merits of large language models become evident.
They not only optimize resource utilization but also outperform smaller models in adaptability, especially in varied application domains and data-limited settings. 
Such strengths position them as primary tools for KG construction and reasoning. 
Yet, while the prowess of LLMs is impressive, researchers have identified certain limitations, such as misalignment with human preferences and the tendency for hallucinations. 
The efficacy of models like ChatGPT heavily leans on human engagement in dialogue generation. 
Further refining model responses necessitates intricate user task descriptions and enriched interaction contexts, a process that remains demanding and time-intensive in the development lifecycle.
\begin{figure*}[!ht]

\centering 
\includegraphics[width=1.0\textwidth]{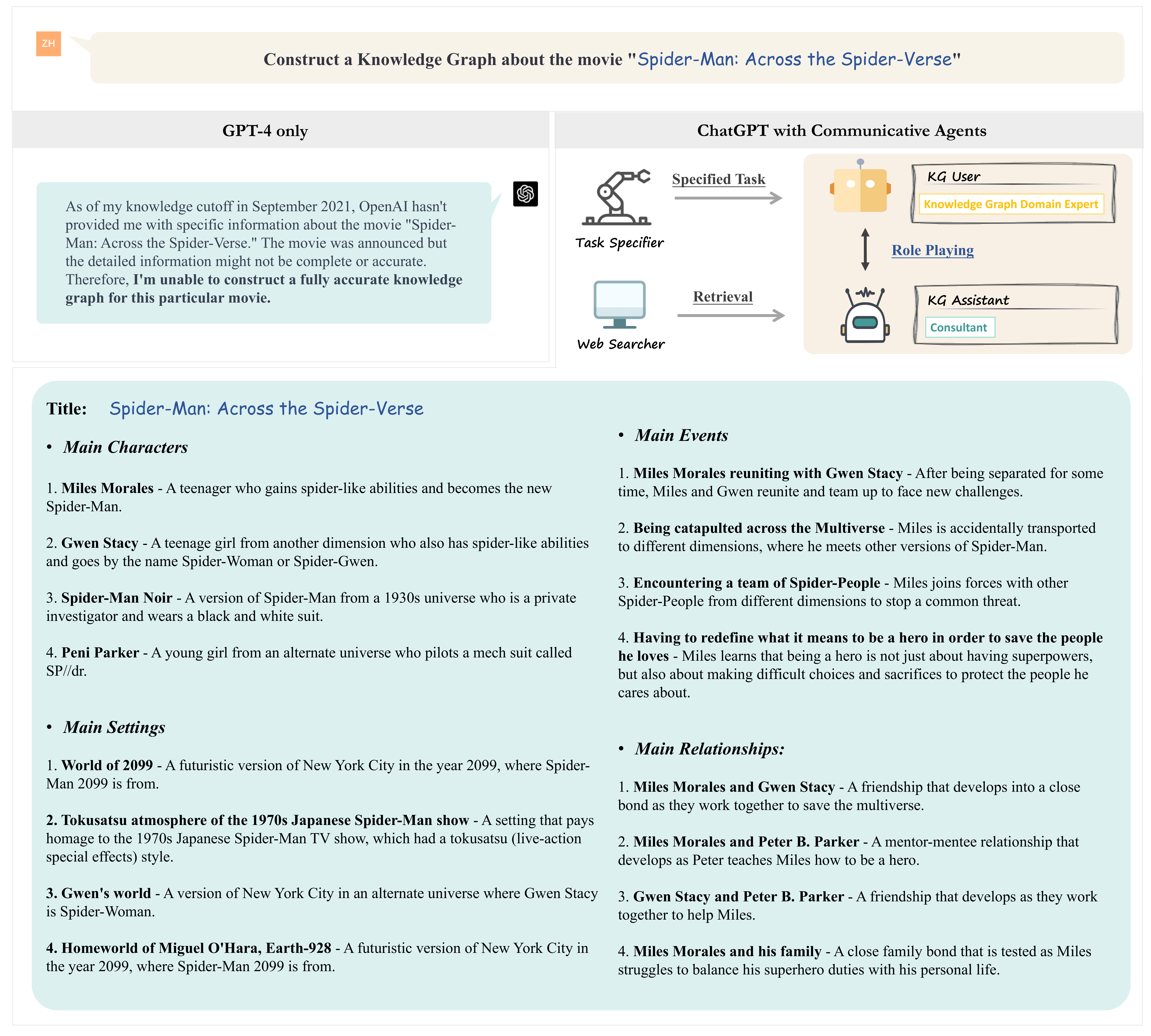} 
\caption{
Illustration of \textbf{AutoKG}, that integrates KG construction and reasoning by employing GPT-4 and communicative agents based on ChatGPT.
The figure omits the specific operational process, providing the results directly.}
\label{fig:KGgraph}
\end{figure*}
Consequently, there is a growing interest in the realm of interactive natural language processing (iNLP)~\cite{CoRR23_INLP}.
In parallel, research efforts concerning intelligent agents continue to proliferate~\cite{A_Survey_on_LLM_based_Autonomous_Agents,The_Rise_and_Potential_of_Agents,An_In_depth_Survey_of_LLM_based_Agents}.
A notable example of this advancement is 
AutoGPT\footnote{\url{https://github.com/Significant-Gravitas/Auto-GPT}}, which can independently generate prompts and carry out tasks such as event analysis, programming, and mathematical operations. Concurrently,~\citet{DBLP:journals/corr/abs-2303-17760} delves into the potential for autonomous cooperation between communicative agents and introduces a novel cooperative agent framework called \textit{role-playing}.

In light of our findings, we propose the use of communicative intelligent agents for KG construction, leveraging different roles assigned to multiple agents to collaborate on KG tasks based on their mutual knowledge. 
Considering the knowledge cutoff prevalent in large models during the pre-training phase, we suggest the incorporation of external sources to assist task completion. 
These sources can include knowledge bases, existing KGs, and internet retrieval systems, among others. 
Here we name this \textbf{AutoKG}.

For a simple demonstration of the concept, we utilize the \textit{role-playing} method in CAMEL~\cite{DBLP:journals/corr/abs-2303-17760}. 
As depicted in Figure~\ref{fig:KGgraph}, we designate the \textit{KG assistant} agent as a \textit{Consultant} and the \textit{KG user} agent as a \textit{KG domain expert}. 
Upon receipt of the prompt and assigned roles, the task-specifier agent provides an elaborate description to clarify the concept. 
Following this, the \textit{KG assistant} and \textit{KG user} collaborate in a multi-party setting to complete the specified task until the \textit{KG user} confirms its completion. 
Concurrently, a \textit{web searcher} role is introduced to aid the \textit{KG assistant} in internet knowledge retrieval.
When the \textit{KG assistant} receives a dialogue from the \textit{KG user}, it initially consults the  \textit{web searcher} on whether to browse information online based on the content. 
Guided by the \textit{web searcher}'s response, the \textit{KG assistant} then continues to address the \textit{KG user}'s command.
The experimental example indicates that the knowledge graph related to the film \textit{Spider-Man: Across the Spider-Verse} released in 2023 is more effectively and comprehensively constructed using the multi-agent and internet-augmented approach. 

\textit{\textbf{Remark.}}
By combining the efforts of artificial intelligence and human expertise, AutoKG could speed up the creation of specialized KGs, fostering a collaborative environment with language models. 
This system leverages domain and internet knowledge to produce high-quality KGs, augmenting the factual accuracy of LLMs in domain-specific tasks, thereby increasing their practical utility. 
AutoKG not only simplifies the construction process but also improves LLMs' transparency, facilitating a deeper understanding of their internal workings. 
As a cooperative human-machine platform, it bolsters the understanding and guidance of LLMs' decision-making, increasing their efficiency in complex tasks. 
However, it is noteworthy that despite the assistance of AutoKG, the current results of the constructed knowledge graph still necessitate manual evaluation and validation.

Furthermore, three significant challenges remain when utilizing AutoKG, necessitating further research and resolution:
~\textbf{The utilization of the API is constrained by a maximum token limit.} 
Currently, 
the gpt-3.5-turbo in use is subjected to a max token restriction. 
This constraint impacts the construction of KGs.
~\textbf{AutoKG now exhibits shortcomings in facilitating efficient human-machine interaction.} 
In fully autonomous machine operations, human oversight for immediate error correction is lacking, yet incorporating human involvement in every step will increase time and labor costs substantially.
~\textbf{Hallucination problem of LLMs.}
Given the known propensity of LLMs to generate non-factual information, it's imperative to scrutinize outputs from them. 
This can be achieved via comparison with standard answers, expert review, or through semi-automatic algorithms.


\section{Conclusion and Future Work}
In this paper, we investigate LLMs for KG construction and reasoning.
We question whether LLMs' extraction abilities arise from their vast pre-training corpus or their strong contextual learning capabilities.
To investigate this, we conduct a Virtual Knowledge Extraction task using a novel dataset, with results highlighting the LLMs' robust contextual learning.
Furthermore, we propose an innovative method of AutoKG for accomplishing KG construction and reasoning tasks by employing multiple agents.
In the future, we would like to extend our work to other LLMs and explore additional KG-related tasks, such as multimodal reasoning.

\section*{Declarations}
\begin{itemize}
\item Funding. 
This work was supported by the National Natural Science Foundation of China (No. 62206246, No. NSFCU23B2055, No. NSFCU19B2027), the Fundamental Research Funds for the Central Universities (226-2023-00138), Zhejiang Provincial Natural Science Foundation of China (No. LGG22F030011), Yongjiang Talent Introduction Programme (2021A-156-G), Tencent AI Lab Rhino-Bird Focused Research Program (RBFR2024003), Information Technology Center and State Key Lab of CAD\&CG, Zhejiang University, and NUS-NCS Joint Laboratory (A-0008542-00-00).
\item Ethics approval and consent to participate. This work did not involve any human participants, their data, or biological materials, and therefore did not require ethical approval. 
\item Data and Materials availability. Our data and materials are accessible in the repository here~\footnote{The code and datasets are in \url{https://github.com/zjunlp/AutoKG}}.
\end{itemize}

\bibliography{sn-bibliography}


\begin{thebibliography}{86}
\ifx \bisbn   \undefined \def \bisbn  #1{ISBN #1}\fi
\ifx \binits  \undefined \def \binits#1{#1}\fi
\ifx \bauthor  \undefined \def \bauthor#1{#1}\fi
\ifx \batitle  \undefined \def \batitle#1{#1}\fi
\ifx \bjtitle  \undefined \def \bjtitle#1{#1}\fi
\ifx \bvolume  \undefined \def \bvolume#1{\textbf{#1}}\fi
\ifx \byear  \undefined \def \byear#1{#1}\fi
\ifx \bissue  \undefined \def \bissue#1{#1}\fi
\ifx \bfpage  \undefined \def \bfpage#1{#1}\fi
\ifx \blpage  \undefined \def \blpage #1{#1}\fi
\ifx \burl  \undefined \def \burl#1{\textsf{#1}}\fi
\ifx \doiurl  \undefined \def \doiurl#1{\url{https://doi.org/#1}}\fi
\ifx \betal  \undefined \def \betal{\textit{et al.}}\fi
\ifx \binstitute  \undefined \def \binstitute#1{#1}\fi
\ifx \binstitutionaled  \undefined \def \binstitutionaled#1{#1}\fi
\ifx \bctitle  \undefined \def \bctitle#1{#1}\fi
\ifx \beditor  \undefined \def \beditor#1{#1}\fi
\ifx \bpublisher  \undefined \def \bpublisher#1{#1}\fi
\ifx \bbtitle  \undefined \def \bbtitle#1{#1}\fi
\ifx \bedition  \undefined \def \bedition#1{#1}\fi
\ifx \bseriesno  \undefined \def \bseriesno#1{#1}\fi
\ifx \blocation  \undefined \def \blocation#1{#1}\fi
\ifx \bsertitle  \undefined \def \bsertitle#1{#1}\fi
\ifx \bsnm \undefined \def \bsnm#1{#1}\fi
\ifx \bsuffix \undefined \def \bsuffix#1{#1}\fi
\ifx \bparticle \undefined \def \bparticle#1{#1}\fi
\ifx \barticle \undefined \def \barticle#1{#1}\fi
\bibcommenthead
\ifx \bconfdate \undefined \def \bconfdate #1{#1}\fi
\ifx \botherref \undefined \def \botherref #1{#1}\fi
\ifx \url \undefined \def \url#1{\textsf{#1}}\fi
\ifx \bchapter \undefined \def \bchapter#1{#1}\fi
\ifx \bbook \undefined \def \bbook#1{#1}\fi
\ifx \bcomment \undefined \def \bcomment#1{#1}\fi
\ifx \oauthor \undefined \def \oauthor#1{#1}\fi
\ifx \citeauthoryear \undefined \def \citeauthoryear#1{#1}\fi
\ifx \endbibitem  \undefined \def \endbibitem {}\fi
\ifx \bconflocation  \undefined \def \bconflocation#1{#1}\fi
\ifx \arxivurl  \undefined \def \arxivurl#1{\textsf{#1}}\fi
\csname PreBibitemsHook\endcsname

\bibitem[\protect\citeauthoryear{Cai et~al.}{2023}]{cai2022temporal}
\begin{bchapter}
\bauthor{\bsnm{Cai}, \binits{B.}},
\bauthor{\bsnm{Xiang}, \binits{Y.}},
\bauthor{\bsnm{Gao}, \binits{L.}},
\bauthor{\bsnm{Zhang}, \binits{H.}},
\bauthor{\bsnm{Li}, \binits{Y.}},
\bauthor{\bsnm{Li}, \binits{J.}}:
\bctitle{Temporal knowledge graph completion: {A} survey}.
In: \bbtitle{Proceedings of the Thirty-Second International Joint Conference on Artificial Intelligence, {IJCAI} 2023, 19th-25th August 2023, Macao, SAR, China},
pp. \bfpage{6545}--\blpage{6553}
(\byear{2023}).
\doiurl{10.24963/IJCAI.2023/734}
\end{bchapter}
\endbibitem

\bibitem[\protect\citeauthoryear{Zhu et~al.}{2024}]{zhu2022multi}
\begin{barticle}
\bauthor{\bsnm{Zhu}, \binits{X.}},
\bauthor{\bsnm{Li}, \binits{Z.}},
\bauthor{\bsnm{Wang}, \binits{X.}},
\bauthor{\bsnm{Jiang}, \binits{X.}},
\bauthor{\bsnm{Sun}, \binits{P.}},
\bauthor{\bsnm{Wang}, \binits{X.}},
\bauthor{\bsnm{Xiao}, \binits{Y.}},
\bauthor{\bsnm{Yuan}, \binits{N.J.}}:
\batitle{Multi-modal knowledge graph construction and application: {A} survey}.
\bjtitle{{IEEE} Trans. Knowl. Data Eng.}
\bvolume{36}(\bissue{2}),
\bfpage{715}--\blpage{735}
(\byear{2024})
\doiurl{10.1109/TKDE.2022.3224228}
\end{barticle}
\endbibitem

\bibitem[\protect\citeauthoryear{Liang et~al.}{2022}]{DBLP:journals/corr/abs-2212-05767}
\begin{botherref}
\oauthor{\bsnm{Liang}, \binits{K.Y.}},
\oauthor{\bsnm{Meng}, \binits{L.}},
\oauthor{\bsnm{Liu}, \binits{M.}},
\oauthor{\bsnm{Liu}, \binits{Y.}},
\oauthor{\bsnm{Tu}, \binits{W.}},
\oauthor{\bsnm{Wang}, \binits{S.}},
\oauthor{\bsnm{Zhou}, \binits{S.}},
\oauthor{\bsnm{Liu}, \binits{X.}},
\oauthor{\bsnm{Sun}, \binits{F.}}:
A survey of knowledge graph reasoning on graph types: Static, dynamic, and multi-modal.
IEEE transactions on pattern analysis and machine intelligence
\textbf{PP}
(2022)
\end{botherref}
\endbibitem

\bibitem[\protect\citeauthoryear{Chen et~al.}{2024}]{DBLP:journals/corr/abs-2305-08698}
\begin{bchapter}
\bauthor{\bsnm{Chen}, \binits{X.}},
\bauthor{\bsnm{Zhang}, \binits{J.}},
\bauthor{\bsnm{Wang}, \binits{X.}},
\bauthor{\bsnm{Wu}, \binits{T.}},
\bauthor{\bsnm{Deng}, \binits{S.}},
\bauthor{\bsnm{Wang}, \binits{Y.}},
\bauthor{\bsnm{Si}, \binits{L.}},
\bauthor{\bsnm{Chen}, \binits{H.}},
\bauthor{\bsnm{Zhang}, \binits{N.}}:
\bctitle{Continual multimodal knowledge graph construction}.
In: \bbtitle{Proceedings of the Thirty-Third International Joint Conference on Artificial Intelligence, {IJCAI} 2024}
(\byear{2024})
\end{bchapter}
\endbibitem

\bibitem[\protect\citeauthoryear{Pan et~al.}{2024}]{Unify_LLM_and_KG}
\begin{barticle}
\bauthor{\bsnm{Pan}, \binits{S.}},
\bauthor{\bsnm{Luo}, \binits{L.}},
\bauthor{\bsnm{Wang}, \binits{Y.}},
\bauthor{\bsnm{Chen}, \binits{C.}},
\bauthor{\bsnm{Wang}, \binits{J.}},
\bauthor{\bsnm{Wu}, \binits{X.}}:
\batitle{Unifying large language models and knowledge graphs: {A} roadmap}.
\bjtitle{{IEEE} Trans. Knowl. Data Eng.}
\bvolume{36}(\bissue{7}),
\bfpage{3580}--\blpage{3599}
(\byear{2024})
\doiurl{10.1109/TKDE.2024.3352100}
\end{barticle}
\endbibitem

\bibitem[\protect\citeauthoryear{Pan et~al.}{2023}]{LLM_and_KG_Opportunities_and_KG}
\begin{barticle}
\bauthor{\bsnm{Pan}, \binits{J.Z.}},
\bauthor{\bsnm{Razniewski}, \binits{S.}},
\bauthor{\bsnm{Kalo}, \binits{J.}},
\bauthor{\bsnm{Singhania}, \binits{S.}},
\bauthor{\bsnm{Chen}, \binits{J.}},
\bauthor{\bsnm{Dietze}, \binits{S.}},
\bauthor{\bsnm{Jabeen}, \binits{H.}},
\bauthor{\bsnm{Omeliyanenko}, \binits{J.}},
\bauthor{\bsnm{Zhang}, \binits{W.}},
\bauthor{\bsnm{Lissandrini}, \binits{M.}},
\bauthor{\bsnm{Biswas}, \binits{R.}},
\bauthor{\bsnm{Melo}, \binits{G.}},
\bauthor{\bsnm{Bonifati}, \binits{A.}},
\bauthor{\bsnm{Vakaj}, \binits{E.}},
\bauthor{\bsnm{Dragoni}, \binits{M.}},
\bauthor{\bsnm{Graux}, \binits{D.}}:
\batitle{Large language models and knowledge graphs: Opportunities and challenges}.
\bjtitle{{TGDK}}
\bvolume{1}(\bissue{1}),
\bfpage{2}--\blpage{1238}
(\byear{2023})
\doiurl{10.4230/.1.1.2}
\end{barticle}
\endbibitem

\bibitem[\protect\citeauthoryear{Ye et~al.}{2022}]{DBLP:conf/emnlp/Ye0CC22}
\begin{bchapter}
\bauthor{\bsnm{Ye}, \binits{H.}},
\bauthor{\bsnm{Zhang}, \binits{N.}},
\bauthor{\bsnm{Chen}, \binits{H.}},
\bauthor{\bsnm{Chen}, \binits{H.}}:
\bctitle{Generative knowledge graph construction: {A} review}.
In: \bbtitle{Proceedings of the 2022 Conference on Empirical Methods in Natural Language Processing, {EMNLP} 2022, Abu Dhabi, United Arab Emirates, December 7-11, 2022},
pp. \bfpage{1}--\blpage{17}
(\byear{2022}).
\burl{https://doi.org/10.18653/v1/2022.emnlp-main.1}
\end{bchapter}
\endbibitem

\bibitem[\protect\citeauthoryear{Ding et~al.}{2024}]{kgc/Theme-specific}
\begin{barticle}
\bauthor{\bsnm{Ding}, \binits{L.}},
\bauthor{\bsnm{Zhou}, \binits{S.}},
\bauthor{\bsnm{Xiao}, \binits{J.}},
\bauthor{\bsnm{Han}, \binits{J.}}:
\batitle{Automated construction of theme-specific knowledge graphs}.
\bjtitle{CoRR}
(\byear{2024})
\doiurl{10.48550/ARXIV.2404.19146}
\end{barticle}
\endbibitem

\bibitem[\protect\citeauthoryear{Chiu and Nichols}{2016}]{DBLP:journals/tacl/ChiuN16}
\begin{barticle}
\bauthor{\bsnm{Chiu}, \binits{J.P.C.}},
\bauthor{\bsnm{Nichols}, \binits{E.}}:
\batitle{Named entity recognition with bidirectional lstm-cnns}.
\bjtitle{Trans. Assoc. Comput. Linguistics}
\bvolume{4},
\bfpage{357}--\blpage{370}
(\byear{2016})
\doiurl{10.1162/tacl\_a\_00104}
\end{barticle}
\endbibitem

\bibitem[\protect\citeauthoryear{Gui et~al.}{2024}]{ie/IEPile}
\begin{bchapter}
\bauthor{\bsnm{Gui}, \binits{H.}},
\bauthor{\bsnm{Yuan}, \binits{L.}},
\bauthor{\bsnm{Ye}, \binits{H.}},
\bauthor{\bsnm{Zhang}, \binits{N.}},
\bauthor{\bsnm{Sun}, \binits{M.}},
\bauthor{\bsnm{Liang}, \binits{L.}},
\bauthor{\bsnm{Chen}, \binits{H.}}:
\bctitle{Iepile: Unearthing large-scale schema-based information extraction corpus}.
In: \bbtitle{Proceedings of the 62nd Annual Meeting of the Association for Computational Linguistics (Volume 2: Short Papers)}
(\byear{2024})
\end{bchapter}
\endbibitem

\bibitem[\protect\citeauthoryear{Zeng et~al.}{2015}]{DBLP:conf/emnlp/ZengLC015}
\begin{bchapter}
\bauthor{\bsnm{Zeng}, \binits{D.}},
\bauthor{\bsnm{Liu}, \binits{K.}},
\bauthor{\bsnm{Chen}, \binits{Y.}},
\bauthor{\bsnm{Zhao}, \binits{J.}}:
\bctitle{Distant supervision for relation extraction via piecewise convolutional neural networks}.
In: \beditor{\bsnm{M{\`{a}}rquez}, \binits{L.}},
\beditor{\bsnm{Callison{-}Burch}, \binits{C.}},
\beditor{\bsnm{Su}, \binits{J.}},
\beditor{\bsnm{Pighin}, \binits{D.}},
\beditor{\bsnm{Marton}, \binits{Y.}} (eds.)
\bbtitle{Proceedings of the 2015 Conference on Empirical Methods in Natural Language Processing, {EMNLP} 2015, Lisbon, Portugal, September 17-21, 2015},
pp. \bfpage{1753}--\blpage{1762}
(\byear{2015}).
\burl{https://doi.org/10.18653/v1/d15-1203}
\end{bchapter}
\endbibitem

\bibitem[\protect\citeauthoryear{Chen et~al.}{2022}]{DBLP:conf/www/ChenZXDYTHSC22}
\begin{bchapter}
\bauthor{\bsnm{Chen}, \binits{X.}},
\bauthor{\bsnm{Zhang}, \binits{N.}},
\bauthor{\bsnm{Xie}, \binits{X.}},
\bauthor{\bsnm{Deng}, \binits{S.}},
\bauthor{\bsnm{Yao}, \binits{Y.}},
\bauthor{\bsnm{Tan}, \binits{C.}},
\bauthor{\bsnm{Huang}, \binits{F.}},
\bauthor{\bsnm{Si}, \binits{L.}},
\bauthor{\bsnm{Chen}, \binits{H.}}:
\bctitle{Knowprompt: Knowledge-aware prompt-tuning with synergistic optimization for relation extraction}.
In: \beditor{\bsnm{Laforest}, \binits{F.}},
\beditor{\bsnm{Troncy}, \binits{R.}},
\beditor{\bsnm{Simperl}, \binits{E.}},
\beditor{\bsnm{Agarwal}, \binits{D.}},
\beditor{\bsnm{Gionis}, \binits{A.}},
\beditor{\bsnm{Herman}, \binits{I.}},
\beditor{\bsnm{M{\'{e}}dini}, \binits{L.}} (eds.)
\bbtitle{{WWW} '22: The {ACM} Web Conference 2022, Virtual Event, Lyon, France, April 25 - 29, 2022},
pp. \bfpage{2778}--\blpage{2788}
(\byear{2022}).
\burl{https://doi.org/10.1145/3485447.3511998}
\end{bchapter}
\endbibitem

\bibitem[\protect\citeauthoryear{Chen et~al.}{2015}]{DBLP:conf/acl/ChenXLZ015}
\begin{bchapter}
\bauthor{\bsnm{Chen}, \binits{Y.}},
\bauthor{\bsnm{Xu}, \binits{L.}},
\bauthor{\bsnm{Liu}, \binits{K.}},
\bauthor{\bsnm{Zeng}, \binits{D.}},
\bauthor{\bsnm{Zhao}, \binits{J.}}:
\bctitle{Event extraction via dynamic multi-pooling convolutional neural networks}.
In: \bbtitle{Proceedings of the 53rd Annual Meeting of the Association for Computational Linguistics and the 7th International Joint Conference on Natural Language Processing of the Asian Federation of Natural Language Processing, {ACL} 2015, July 26-31, 2015, Beijing, China, Volume 1: Long Papers},
pp. \bfpage{167}--\blpage{176}
(\byear{2015}).
\burl{https://doi.org/10.3115/v1/p15-1017}
\end{bchapter}
\endbibitem

\bibitem[\protect\citeauthoryear{Deng et~al.}{2020}]{DBLP:conf/wsdm/DengZKZZC20}
\begin{bchapter}
\bauthor{\bsnm{Deng}, \binits{S.}},
\bauthor{\bsnm{Zhang}, \binits{N.}},
\bauthor{\bsnm{Kang}, \binits{J.}},
\bauthor{\bsnm{Zhang}, \binits{Y.}},
\bauthor{\bsnm{Zhang}, \binits{W.}},
\bauthor{\bsnm{Chen}, \binits{H.}}:
\bctitle{Meta-learning with dynamic-memory-based prototypical network for few-shot event detection}.
In: \beditor{\bsnm{Caverlee}, \binits{J.}},
\beditor{\bsnm{Hu}, \binits{X.B.}},
\beditor{\bsnm{Lalmas}, \binits{M.}},
\beditor{\bsnm{Wang}, \binits{W.}} (eds.)
\bbtitle{{WSDM} '20: The Thirteenth {ACM} International Conference on Web Search and Data Mining, Houston, TX, USA, February 3-7, 2020},
pp. \bfpage{151}--\blpage{159}
(\byear{2020}).
\burl{https://doi.org/10.1145/3336191.3371796}
\end{bchapter}
\endbibitem

\bibitem[\protect\citeauthoryear{Shen et~al.}{2015}]{DBLP:journals/tkde/ShenWH15}
\begin{barticle}
\bauthor{\bsnm{Shen}, \binits{W.}},
\bauthor{\bsnm{Wang}, \binits{J.}},
\bauthor{\bsnm{Han}, \binits{J.}}:
\batitle{Entity linking with a knowledge base: Issues, techniques, and solutions}.
\bjtitle{{IEEE} Trans. Knowl. Data Eng.}
\bvolume{27}(\bissue{2}),
\bfpage{443}--\blpage{460}
(\byear{2015})
\doiurl{10.1109/TKDE.2014.2327028}
\end{barticle}
\endbibitem

\bibitem[\protect\citeauthoryear{Zhang et~al.}{2018}]{DBLP:conf/aaai/ZhangDKSS18}
\begin{bchapter}
\bauthor{\bsnm{Zhang}, \binits{Y.}},
\bauthor{\bsnm{Dai}, \binits{H.}},
\bauthor{\bsnm{Kozareva}, \binits{Z.}},
\bauthor{\bsnm{Smola}, \binits{A.J.}},
\bauthor{\bsnm{Song}, \binits{L.}}:
\bctitle{Variational reasoning for question answering with knowledge graph}.
In: \beditor{\bsnm{McIlraith}, \binits{S.A.}},
\beditor{\bsnm{Weinberger}, \binits{K.Q.}} (eds.)
\bbtitle{Proceedings of the Thirty-Second {AAAI} Conference on Artificial Intelligence, (AAAI-18), the 30th Innovative Applications of Artificial Intelligence (IAAI-18), and the 8th {AAAI} Symposium on Educational Advances in Artificial Intelligence (EAAI-18), New Orleans, Louisiana, USA, February 2-7, 2018}
(\byear{2018}).
\burl{https://doi.org/10.1609/aaai.v32i1.12057}
\end{bchapter}
\endbibitem

\bibitem[\protect\citeauthoryear{Rossi et~al.}{2021}]{DBLP:journals/tkdd/RossiBFMM21}
\begin{barticle}
\bauthor{\bsnm{Rossi}, \binits{A.}},
\bauthor{\bsnm{Barbosa}, \binits{D.}},
\bauthor{\bsnm{Firmani}, \binits{D.}},
\bauthor{\bsnm{Matinata}, \binits{A.}},
\bauthor{\bsnm{Merialdo}, \binits{P.}}:
\batitle{Knowledge graph embedding for link prediction: {A} comparative analysis}.
\bjtitle{{ACM} Trans. Knowl. Discov. Data}
\bvolume{15}(\bissue{2}),
\bfpage{14}--\blpage{11449}
(\byear{2021})
\doiurl{10.1145/3424672}
\end{barticle}
\endbibitem

\bibitem[\protect\citeauthoryear{Karpukhin et~al.}{2020}]{DBLP:conf/emnlp/KarpukhinOMLWEC20}
\begin{bchapter}
\bauthor{\bsnm{Karpukhin}, \binits{V.}},
\bauthor{\bsnm{Oguz}, \binits{B.}},
\bauthor{\bsnm{Min}, \binits{S.}},
\bauthor{\bsnm{Lewis}, \binits{P.S.H.}},
\bauthor{\bsnm{Wu}, \binits{L.}},
\bauthor{\bsnm{Edunov}, \binits{S.}},
\bauthor{\bsnm{Chen}, \binits{D.}},
\bauthor{\bsnm{Yih}, \binits{W.}}:
\bctitle{Dense passage retrieval for open-domain question answering}.
In: \beditor{\bsnm{Webber}, \binits{B.}},
\beditor{\bsnm{Cohn}, \binits{T.}},
\beditor{\bsnm{He}, \binits{Y.}},
\beditor{\bsnm{Liu}, \binits{Y.}} (eds.)
\bbtitle{Proceedings of the 2020 Conference on Empirical Methods in Natural Language Processing, {EMNLP} 2020, Online, November 16-20, 2020}
(\byear{2020}).
\burl{https://doi.org/10.18653/v1/2020.emnlp-main.550}
\end{bchapter}
\endbibitem

\bibitem[\protect\citeauthoryear{Zhu et~al.}{2021}]{DBLP:journals/corr/abs-2101-00774}
\begin{botherref}
\oauthor{\bsnm{Zhu}, \binits{F.}},
\oauthor{\bsnm{Lei}, \binits{W.}},
\oauthor{\bsnm{Wang}, \binits{C.}},
\oauthor{\bsnm{Zheng}, \binits{J.}},
\oauthor{\bsnm{Poria}, \binits{S.}},
\oauthor{\bsnm{Chua}, \binits{T.}}:
Retrieving and reading: {A} comprehensive survey on open-domain question answering.
CoRR
(2021)
\end{botherref}
\endbibitem

\bibitem[\protect\citeauthoryear{OpenAI}{2023}]{DBLP:journals/corr/abs-2303-08774}
\begin{botherref}
\oauthor{\bsnm{OpenAI}}:
{GPT-4} technical report.
CoRR
\textbf{abs/2303.08774}
(2023)
\doiurl{10.48550/arXiv.2303.08774}
\end{botherref}
\endbibitem

\bibitem[\protect\citeauthoryear{Liu et~al.}{2023}]{DBLP:journals/corr/abs-2303-13547}
\begin{barticle}
\bauthor{\bsnm{Liu}, \binits{A.}},
\bauthor{\bsnm{Hu}, \binits{X.}},
\bauthor{\bsnm{Wen}, \binits{L.}},
\bauthor{\bsnm{Yu}, \binits{P.S.}}:
\batitle{A comprehensive evaluation of chatgpt's zero-shot text-to-sql capability}.
\bjtitle{CoRR}
(\byear{2023})
\doiurl{10.48550/arXiv.2303.13547}
\end{barticle}
\endbibitem

\bibitem[\protect\citeauthoryear{Shakarian et~al.}{2023}]{DBLP:journals/corr/abs-2302-13814}
\begin{bchapter}
\bauthor{\bsnm{Shakarian}, \binits{P.}},
\bauthor{\bsnm{Koyyalamudi}, \binits{A.}},
\bauthor{\bsnm{Ngu}, \binits{N.}},
\bauthor{\bsnm{Mareedu}, \binits{L.}}:
\bctitle{An independent evaluation of chatgpt on mathematical word problems {(MWP)}}.
In: \beditor{\bsnm{Martin}, \binits{A.}},
\beditor{\bsnm{Fill}, \binits{H.}},
\beditor{\bsnm{Gerber}, \binits{A.}},
\beditor{\bsnm{Hinkelmann}, \binits{K.}},
\beditor{\bsnm{Lenat}, \binits{D.}},
\beditor{\bsnm{Stolle}, \binits{R.}},
\beditor{\bsnm{Harmelen}, \binits{F.}} (eds.)
\bbtitle{Proceedings of the {AAAI} 2023 Spring Symposium on Challenges Requiring the Combination of Machine Learning and Knowledge Engineering {(AAAI-MAKE} 2023), Hyatt Regency, San Francisco Airport, California, USA, March 27-29, 2023}.
\bsertitle{{CEUR} Workshop Proceedings},
vol. \bseriesno{3433}
(\byear{2023})
\end{bchapter}
\endbibitem

\bibitem[\protect\citeauthoryear{Lai et~al.}{2023}]{DBLP:journals/corr/abs-2304-05613}
\begin{bchapter}
\bauthor{\bsnm{Lai}, \binits{V.D.}},
\bauthor{\bsnm{Ngo}, \binits{N.T.}},
\bauthor{\bsnm{Veyseh}, \binits{A.P.B.}},
\bauthor{\bsnm{Man}, \binits{H.}},
\bauthor{\bsnm{Dernoncourt}, \binits{F.}},
\bauthor{\bsnm{Bui}, \binits{T.}},
\bauthor{\bsnm{Nguyen}, \binits{T.H.}}:
\bctitle{Chatgpt beyond english: Towards a comprehensive evaluation of large language models in multilingual learning}.
In: \beditor{\bsnm{Bouamor}, \binits{H.}},
\beditor{\bsnm{Pino}, \binits{J.}},
\beditor{\bsnm{Bali}, \binits{K.}} (eds.)
\bbtitle{Findings of the Association for Computational Linguistics: {EMNLP} 2023, Singapore, December 6-10, 2023}
(\byear{2023}).
\burl{https://doi.org/10.18653/v1/2023.findings-emnlp.878}
\end{bchapter}
\endbibitem

\bibitem[\protect\citeauthoryear{Zhao et~al.}{2023}]{DBLP:journals/corr/abs-2303-18223}
\begin{barticle}
\bauthor{\bsnm{Zhao}, \binits{W.X.}},
\bauthor{\bsnm{Zhou}, \binits{K.}},
\bauthor{\bsnm{Li}, \binits{J.}},
\bauthor{\bsnm{Tang}, \binits{T.}},
\bauthor{\bsnm{Wang}, \binits{X.}},
\bauthor{\bsnm{Hou}, \binits{Y.}},
\bauthor{\bsnm{Min}, \binits{Y.}},
\bauthor{\bsnm{Zhang}, \binits{B.}},
\bauthor{\bsnm{Zhang}, \binits{J.}},
\bauthor{\bsnm{Dong}, \binits{Z.}},
\bauthor{\bsnm{Du}, \binits{Y.}},
\bauthor{\bsnm{Yang}, \binits{C.}},
\bauthor{\bsnm{Chen}, \binits{Y.}},
\bauthor{\bsnm{Chen}, \binits{Z.}},
\bauthor{\bsnm{Jiang}, \binits{J.}},
\bauthor{\bsnm{Ren}, \binits{R.}},
\bauthor{\bsnm{Li}, \binits{Y.}},
\bauthor{\bsnm{Tang}, \binits{X.}},
\bauthor{\bsnm{Liu}, \binits{Z.}},
\bauthor{\bsnm{Liu}, \binits{P.}},
\bauthor{\bsnm{Nie}, \binits{J.}},
\bauthor{\bsnm{Wen}, \binits{J.}}:
\batitle{A survey of large language models}.
\bjtitle{CoRR}
(\byear{2023})
\doiurl{10.48550/arXiv.2303.18223}
\end{barticle}
\endbibitem

\bibitem[\protect\citeauthoryear{Wei et~al.}{2023}]{DBLP:journals/corr/abs-2302-10205}
\begin{botherref}
\oauthor{\bsnm{Wei}, \binits{X.}},
\oauthor{\bsnm{Cui}, \binits{X.}},
\oauthor{\bsnm{Cheng}, \binits{N.}},
\oauthor{\bsnm{Wang}, \binits{X.}},
\oauthor{\bsnm{Zhang}, \binits{X.}},
\oauthor{\bsnm{Huang}, \binits{S.}},
\oauthor{\bsnm{Xie}, \binits{P.}},
\oauthor{\bsnm{Xu}, \binits{J.}},
\oauthor{\bsnm{Chen}, \binits{Y.}},
\oauthor{\bsnm{Zhang}, \binits{M.}},
\oauthor{\bsnm{Jiang}, \binits{Y.}},
\oauthor{\bsnm{Han}, \binits{W.}}:
Zero-shot information extraction via chatting with chatgpt.
CoRR
\textbf{abs/2302.10205}
(2023)
\doiurl{10.48550/arXiv.2302.10205}
\end{botherref}
\endbibitem

\bibitem[\protect\citeauthoryear{Li et~al.}{2023a}]{eval/Capabilities}
\begin{botherref}
\oauthor{\bsnm{Li}, \binits{B.}},
\oauthor{\bsnm{Fang}, \binits{G.}},
\oauthor{\bsnm{Yang}, \binits{Y.}},
\oauthor{\bsnm{Wang}, \binits{Q.}},
\oauthor{\bsnm{Ye}, \binits{W.}},
\oauthor{\bsnm{Zhao}, \binits{W.}},
\oauthor{\bsnm{Zhang}, \binits{S.}}:
Evaluating chatgpt's information extraction capabilities: An assessment of performance, explainability, calibration, and faithfulness.
CoRR
(2023)
\end{botherref}
\endbibitem

\bibitem[\protect\citeauthoryear{Li et~al.}{2023b}]{DBLP:conf/emnlp/LiWK23}
\begin{bchapter}
\bauthor{\bsnm{Li}, \binits{G.}},
\bauthor{\bsnm{Wang}, \binits{P.}},
\bauthor{\bsnm{Ke}, \binits{W.}}:
\bctitle{Revisiting large language models as zero-shot relation extractors}.
In: \beditor{\bsnm{Bouamor}, \binits{H.}},
\beditor{\bsnm{Pino}, \binits{J.}},
\beditor{\bsnm{Bali}, \binits{K.}} (eds.)
\bbtitle{Findings of the Association for Computational Linguistics: {EMNLP} 2023, Singapore, December 6-10, 2023}
(\byear{2023}).
\burl{https://doi.org/10.18653/v1/2023.findings-emnlp.459}
\end{bchapter}
\endbibitem

\bibitem[\protect\citeauthoryear{Wan et~al.}{2023}]{DBLP:conf/emnlp/WanCMLSLK23}
\begin{bchapter}
\bauthor{\bsnm{Wan}, \binits{Z.}},
\bauthor{\bsnm{Cheng}, \binits{F.}},
\bauthor{\bsnm{Mao}, \binits{Z.}},
\bauthor{\bsnm{Liu}, \binits{Q.}},
\bauthor{\bsnm{Song}, \binits{H.}},
\bauthor{\bsnm{Li}, \binits{J.}},
\bauthor{\bsnm{Kurohashi}, \binits{S.}}:
\bctitle{{GPT-RE:} in-context learning for relation extraction using large language models}.
In: \beditor{\bsnm{Bouamor}, \binits{H.}},
\beditor{\bsnm{Pino}, \binits{J.}},
\beditor{\bsnm{Bali}, \binits{K.}} (eds.)
\bbtitle{Proceedings of the 2023 Conference on Empirical Methods in Natural Language Processing, {EMNLP} 2023, Singapore, December 6-10, 2023}
(\byear{2023}).
\burl{https://doi.org/10.18653/v1/2023.emnlp-main.214}
\end{bchapter}
\endbibitem

\bibitem[\protect\citeauthoryear{Qin et~al.}{2023}]{reason/nlptasksolver}
\begin{bchapter}
\bauthor{\bsnm{Qin}, \binits{C.}},
\bauthor{\bsnm{Zhang}, \binits{A.}},
\bauthor{\bsnm{Zhang}, \binits{Z.}},
\bauthor{\bsnm{Chen}, \binits{J.}},
\bauthor{\bsnm{Yasunaga}, \binits{M.}},
\bauthor{\bsnm{Yang}, \binits{D.}}:
\bctitle{Is chatgpt a general-purpose natural language processing task solver?}
In: \beditor{\bsnm{Bouamor}, \binits{H.}},
\beditor{\bsnm{Pino}, \binits{J.}},
\beditor{\bsnm{Bali}, \binits{K.}} (eds.)
\bbtitle{Proceedings of the 2023 Conference on Empirical Methods in Natural Language Processing, {EMNLP} 2023, Singapore, December 6-10, 2023}
(\byear{2023}).
\burl{https://doi.org/10.18653/v1/2023.emnlp-main.85}
\end{bchapter}
\endbibitem

\bibitem[\protect\citeauthoryear{Liu et~al.}{2023}]{reason/logicalreasoning}
\begin{barticle}
\bauthor{\bsnm{Liu}, \binits{H.}},
\bauthor{\bsnm{Ning}, \binits{R.}},
\bauthor{\bsnm{Teng}, \binits{Z.}},
\bauthor{\bsnm{Liu}, \binits{J.}},
\bauthor{\bsnm{Zhou}, \binits{Q.}},
\bauthor{\bsnm{Zhang}, \binits{Y.}}:
\batitle{Evaluating the logical reasoning ability of chatgpt and {GPT-4}}.
\bjtitle{CoRR}
(\byear{2023})
\doiurl{10.48550/ARXIV.2304.03439}
\end{barticle}
\endbibitem

\bibitem[\protect\citeauthoryear{Jiang et~al.}{2024}]{kgr/kgagent}
\begin{barticle}
\bauthor{\bsnm{Jiang}, \binits{J.}},
\bauthor{\bsnm{Zhou}, \binits{K.}},
\bauthor{\bsnm{Zhao}, \binits{W.X.}},
\bauthor{\bsnm{Song}, \binits{Y.}},
\bauthor{\bsnm{Zhu}, \binits{C.}},
\bauthor{\bsnm{Zhu}, \binits{H.}},
\bauthor{\bsnm{Wen}, \binits{J.}}:
\batitle{Kg-agent: An efficient autonomous agent framework for complex reasoning over knowledge graph}.
\bjtitle{CoRR}
(\byear{2024})
\doiurl{10.48550/ARXIV.2402.11163}
\end{barticle}
\endbibitem

\bibitem[\protect\citeauthoryear{Longpre et~al.}{2023}]{thenflancollection}
\begin{bchapter}
\bauthor{\bsnm{Longpre}, \binits{S.}},
\bauthor{\bsnm{Hou}, \binits{L.}},
\bauthor{\bsnm{Vu}, \binits{T.}},
\bauthor{\bsnm{Webson}, \binits{A.}},
\bauthor{\bsnm{Chung}, \binits{H.W.}},
\bauthor{\bsnm{Tay}, \binits{Y.}},
\bauthor{\bsnm{Zhou}, \binits{D.}},
\bauthor{\bsnm{Le}, \binits{Q.V.}},
\bauthor{\bsnm{Zoph}, \binits{B.}},
\bauthor{\bsnm{Wei}, \binits{J.}},
\bauthor{\bsnm{Roberts}, \binits{A.}}:
\bctitle{The flan collection: Designing data and methods for effective instruction tuning}.
In: \beditor{\bsnm{Krause}, \binits{A.}},
\beditor{\bsnm{Brunskill}, \binits{E.}},
\beditor{\bsnm{Cho}, \binits{K.}},
\beditor{\bsnm{Engelhardt}, \binits{B.}},
\beditor{\bsnm{Sabato}, \binits{S.}},
\beditor{\bsnm{Scarlett}, \binits{J.}} (eds.)
\bbtitle{International Conference on Machine Learning, {ICML} 2023, 23-29 July 2023, Honolulu, Hawaii, {USA}}.
\bsertitle{Proceedings of Machine Learning Research}
(\byear{2023})
\end{bchapter}
\endbibitem

\bibitem[\protect\citeauthoryear{Christiano et~al.}{2017}]{DBLP:conf/nips/ChristianoLBMLA17}
\begin{bchapter}
\bauthor{\bsnm{Christiano}, \binits{P.F.}},
\bauthor{\bsnm{Leike}, \binits{J.}},
\bauthor{\bsnm{Brown}, \binits{T.B.}},
\bauthor{\bsnm{Martic}, \binits{M.}},
\bauthor{\bsnm{Legg}, \binits{S.}},
\bauthor{\bsnm{Amodei}, \binits{D.}}:
\bctitle{Deep reinforcement learning from human preferences}.
In: \beditor{\bsnm{Guyon}, \binits{I.}},
\beditor{\bsnm{Luxburg}, \binits{U.}},
\beditor{\bsnm{Bengio}, \binits{S.}},
\beditor{\bsnm{Wallach}, \binits{H.M.}},
\beditor{\bsnm{Fergus}, \binits{R.}},
\beditor{\bsnm{Vishwanathan}, \binits{S.V.N.}},
\beditor{\bsnm{Garnett}, \binits{R.}} (eds.)
\bbtitle{Advances in Neural Information Processing Systems 30: Annual Conference on Neural Information Processing Systems 2017, December 4-9, 2017, Long Beach, CA, {USA}}
(\byear{2017})
\end{bchapter}
\endbibitem

\bibitem[\protect\citeauthoryear{Leiter et~al.}{2023}]{chatgpt_analysis}
\begin{barticle}
\bauthor{\bsnm{Leiter}, \binits{C.}},
\bauthor{\bsnm{Zhang}, \binits{R.}},
\bauthor{\bsnm{Chen}, \binits{Y.}},
\bauthor{\bsnm{Belouadi}, \binits{J.}},
\bauthor{\bsnm{Larionov}, \binits{D.}},
\bauthor{\bsnm{Fresen}, \binits{V.}},
\bauthor{\bsnm{Eger}, \binits{S.}}:
\batitle{Chatgpt: {A} meta-analysis after 2.5 months}.
\bjtitle{CoRR}
(\byear{2023})
\doiurl{10.48550/ARXIV.2302.13795}
\end{barticle}
\endbibitem

\bibitem[\protect\citeauthoryear{Yang et~al.}{2024}]{survey/harness_power_of_llm}
\begin{barticle}
\bauthor{\bsnm{Yang}, \binits{J.}},
\bauthor{\bsnm{Jin}, \binits{H.}},
\bauthor{\bsnm{Tang}, \binits{R.}},
\bauthor{\bsnm{Han}, \binits{X.}},
\bauthor{\bsnm{Feng}, \binits{Q.}},
\bauthor{\bsnm{Jiang}, \binits{H.}},
\bauthor{\bsnm{Zhong}, \binits{S.}},
\bauthor{\bsnm{Yin}, \binits{B.}},
\bauthor{\bsnm{Hu}, \binits{X.B.}}:
\batitle{Harnessing the power of llms in practice: {A} survey on chatgpt and beyond}.
\bjtitle{{ACM} Trans. Knowl. Discov. Data}
\bvolume{18}(\bissue{6}),
\bfpage{160}--\blpage{116032}
(\byear{2024})
\doiurl{10.1145/3649506}
\end{barticle}
\endbibitem

\bibitem[\protect\citeauthoryear{Wei et~al.}{2022}]{cot}
\begin{bchapter}
\bauthor{\bsnm{Wei}, \binits{J.}},
\bauthor{\bsnm{Wang}, \binits{X.}},
\bauthor{\bsnm{Schuurmans}, \binits{D.}},
\bauthor{\bsnm{Bosma}, \binits{M.}},
\bauthor{\bsnm{Ichter}, \binits{B.}},
\bauthor{\bsnm{Xia}, \binits{F.}},
\bauthor{\bsnm{Chi}, \binits{E.H.}},
\bauthor{\bsnm{Le}, \binits{Q.V.}},
\bauthor{\bsnm{Zhou}, \binits{D.}}:
\bctitle{Chain-of-thought prompting elicits reasoning in large language models}.
In: \beditor{\bsnm{Koyejo}, \binits{S.}},
\beditor{\bsnm{Mohamed}, \binits{S.}},
\beditor{\bsnm{Agarwal}, \binits{A.}},
\beditor{\bsnm{Belgrave}, \binits{D.}},
\beditor{\bsnm{Cho}, \binits{K.}},
\beditor{\bsnm{Oh}, \binits{A.}} (eds.)
\bbtitle{Advances in Neural Information Processing Systems 35: Annual Conference on Neural Information Processing Systems 2022, NeurIPS 2022, New Orleans, LA, USA, November 28 - December 9, 2022}
(\byear{2022})
\end{bchapter}
\endbibitem

\bibitem[\protect\citeauthoryear{Wang et~al.}{2023}]{CoRR23_INLP}
\begin{barticle}
\bauthor{\bsnm{Wang}, \binits{Z.}},
\bauthor{\bsnm{Zhang}, \binits{G.}},
\bauthor{\bsnm{Yang}, \binits{K.}},
\bauthor{\bsnm{Shi}, \binits{N.}},
\bauthor{\bsnm{Zhou}, \binits{W.}},
\bauthor{\bsnm{Hao}, \binits{S.}},
\bauthor{\bsnm{Xiong}, \binits{G.}},
\bauthor{\bsnm{Li}, \binits{Y.}},
\bauthor{\bsnm{Sim}, \binits{M.Y.}},
\bauthor{\bsnm{Chen}, \binits{X.}},
\bauthor{\bsnm{Zhu}, \binits{Q.}},
\bauthor{\bsnm{Yang}, \binits{Z.}},
\bauthor{\bsnm{Nik}, \binits{A.}},
\bauthor{\bsnm{Liu}, \binits{Q.}},
\bauthor{\bsnm{Lin}, \binits{C.}},
\bauthor{\bsnm{Wang}, \binits{S.}},
\bauthor{\bsnm{Liu}, \binits{R.}},
\bauthor{\bsnm{Chen}, \binits{W.}},
\bauthor{\bsnm{Xu}, \binits{K.}},
\bauthor{\bsnm{Liu}, \binits{D.}},
\bauthor{\bsnm{Guo}, \binits{Y.}},
\bauthor{\bsnm{Fu}, \binits{J.}}:
\batitle{Interactive natural language processing}.
\bjtitle{CoRR}
(\byear{2023})
\doiurl{10.48550/arXiv.2305.13246}
\end{barticle}
\endbibitem

\bibitem[\protect\citeauthoryear{Bubeck et~al.}{2023}]{bubeck2023sparks}
\begin{barticle}
\bauthor{\bsnm{Bubeck}, \binits{S.}},
\bauthor{\bsnm{Chandrasekaran}, \binits{V.}},
\bauthor{\bsnm{Eldan}, \binits{R.}},
\bauthor{\bsnm{Gehrke}, \binits{J.}},
\bauthor{\bsnm{Horvitz}, \binits{E.}},
\bauthor{\bsnm{Kamar}, \binits{E.}},
\bauthor{\bsnm{Lee}, \binits{P.}},
\bauthor{\bsnm{Lee}, \binits{Y.T.}},
\bauthor{\bsnm{Li}, \binits{Y.}},
\bauthor{\bsnm{Lundberg}, \binits{S.M.}},
\bauthor{\bsnm{Nori}, \binits{H.}},
\bauthor{\bsnm{Palangi}, \binits{H.}},
\bauthor{\bsnm{Ribeiro}, \binits{M.T.}},
\bauthor{\bsnm{Zhang}, \binits{Y.}}:
\batitle{Sparks of artificial general intelligence: Early experiments with {GPT-4}}.
\bjtitle{CoRR}
(\byear{2023})
\doiurl{10.48550/ARXIV.2303.12712}
\end{barticle}
\endbibitem

\bibitem[\protect\citeauthoryear{Li et~al.}{2019}]{DBLP:conf/nlpcc/LiHSJLJZLZ19}
\begin{bchapter}
\bauthor{\bsnm{Li}, \binits{S.}},
\bauthor{\bsnm{He}, \binits{W.}},
\bauthor{\bsnm{Shi}, \binits{Y.}},
\bauthor{\bsnm{Jiang}, \binits{W.}},
\bauthor{\bsnm{Liang}, \binits{H.}},
\bauthor{\bsnm{Jiang}, \binits{Y.}},
\bauthor{\bsnm{Zhang}, \binits{Y.}},
\bauthor{\bsnm{Lyu}, \binits{Y.}},
\bauthor{\bsnm{Zhu}, \binits{Y.}}:
\bctitle{Duie: {A} large-scale chinese dataset for information extraction}.
In: \beditor{\bsnm{Tang}, \binits{J.}},
\beditor{\bsnm{Kan}, \binits{M.}},
\beditor{\bsnm{Zhao}, \binits{D.}},
\beditor{\bsnm{Li}, \binits{S.}},
\beditor{\bsnm{Zan}, \binits{H.}} (eds.)
\bbtitle{Natural Language Processing and Chinese Computing - 8th {CCF} International Conference, {NLPCC} 2019, Dunhuang, China, October 9-14, 2019, Proceedings, Part {II}}.
\bsertitle{Lecture Notes in Computer Science},
vol. \bseriesno{11839},
pp. \bfpage{791}--\blpage{800}
(\byear{2019}).
\burl{https://doi.org/10.1007/978-3-030-32236-6\_72}
\end{bchapter}
\endbibitem

\bibitem[\protect\citeauthoryear{Luan et~al.}{2018}]{DBLP:conf/emnlp/LuanHOH18}
\begin{bchapter}
\bauthor{\bsnm{Luan}, \binits{Y.}},
\bauthor{\bsnm{He}, \binits{L.}},
\bauthor{\bsnm{Ostendorf}, \binits{M.}},
\bauthor{\bsnm{Hajishirzi}, \binits{H.}}:
\bctitle{Multi-task identification of entities, relations, and coreference for scientific knowledge graph construction}.
In: \beditor{\bsnm{Riloff}, \binits{E.}},
\beditor{\bsnm{Chiang}, \binits{D.}},
\beditor{\bsnm{Hockenmaier}, \binits{J.}},
\beditor{\bsnm{Tsujii}, \binits{J.}} (eds.)
\bbtitle{Proceedings of the 2018 Conference on Empirical Methods in Natural Language Processing, Brussels, Belgium, October 31 - November 4, 2018},
pp. \bfpage{3219}--\blpage{3232}
(\byear{2018}).
\burl{https://doi.org/10.18653/v1/d18-1360}
\end{bchapter}
\endbibitem

\bibitem[\protect\citeauthoryear{Stoica et~al.}{2021}]{DBLP:conf/aaai/StoicaPP21}
\begin{bchapter}
\bauthor{\bsnm{Stoica}, \binits{G.}},
\bauthor{\bsnm{Platanios}, \binits{E.A.}},
\bauthor{\bsnm{P{\'{o}}czos}, \binits{B.}}:
\bctitle{Re-tacred: Addressing shortcomings of the {TACRED} dataset}.
In: \bbtitle{Thirty-Fifth {AAAI} Conference on Artificial Intelligence, {AAAI} 2021, Thirty-Third Conference on Innovative Applications of Artificial Intelligence, {IAAI} 2021, The Eleventh Symposium on Educational Advances in Artificial Intelligence, {EAAI} 2021, Virtual Event, February 2-9, 2021},
pp. \bfpage{13843}--\blpage{13850}
(\byear{2021})
\end{bchapter}
\endbibitem

\bibitem[\protect\citeauthoryear{Wang et~al.}{2020}]{DBLP:conf/emnlp/WangWHJHLLLLZ20}
\begin{bchapter}
\bauthor{\bsnm{Wang}, \binits{X.}},
\bauthor{\bsnm{Wang}, \binits{Z.}},
\bauthor{\bsnm{Han}, \binits{X.}},
\bauthor{\bsnm{Jiang}, \binits{W.}},
\bauthor{\bsnm{Han}, \binits{R.}},
\bauthor{\bsnm{Liu}, \binits{Z.}},
\bauthor{\bsnm{Li}, \binits{J.}},
\bauthor{\bsnm{Li}, \binits{P.}},
\bauthor{\bsnm{Lin}, \binits{Y.}},
\bauthor{\bsnm{Zhou}, \binits{J.}}:
\bctitle{{MAVEN:} {A} massive general domain event detection dataset}.
In: \beditor{\bsnm{Webber}, \binits{B.}},
\beditor{\bsnm{Cohn}, \binits{T.}},
\beditor{\bsnm{He}, \binits{Y.}},
\beditor{\bsnm{Liu}, \binits{Y.}} (eds.)
\bbtitle{Proceedings of the 2020 Conference on Empirical Methods in Natural Language Processing, {EMNLP} 2020, Online, November 16-20, 2020},
pp. \bfpage{1652}--\blpage{1671}
(\byear{2020}).
\burl{https://doi.org/10.18653/v1/2020.emnlp-main.129}
\end{bchapter}
\endbibitem

\bibitem[\protect\citeauthoryear{Toutanova et~al.}{2015}]{DBLP:conf/emnlp/ToutanovaCPPCG15}
\begin{bchapter}
\bauthor{\bsnm{Toutanova}, \binits{K.}},
\bauthor{\bsnm{Chen}, \binits{D.}},
\bauthor{\bsnm{Pantel}, \binits{P.}},
\bauthor{\bsnm{Poon}, \binits{H.}},
\bauthor{\bsnm{Choudhury}, \binits{P.}},
\bauthor{\bsnm{Gamon}, \binits{M.}}:
\bctitle{Representing text for joint embedding of text and knowledge bases}.
In: \beditor{\bsnm{M{\`{a}}rquez}, \binits{L.}},
\beditor{\bsnm{Callison{-}Burch}, \binits{C.}},
\beditor{\bsnm{Su}, \binits{J.}},
\beditor{\bsnm{Pighin}, \binits{D.}},
\beditor{\bsnm{Marton}, \binits{Y.}} (eds.)
\bbtitle{Proceedings of the 2015 Conference on Empirical Methods in Natural Language Processing, {EMNLP} 2015, Lisbon, Portugal, September 17-21, 2015},
pp. \bfpage{1499}--\blpage{1509}
(\byear{2015}).
\burl{https://doi.org/10.18653/v1/d15-1174}
\end{bchapter}
\endbibitem

\bibitem[\protect\citeauthoryear{Hwang et~al.}{2021}]{DBLP:conf/aaai/HwangBBDSBC21}
\begin{bchapter}
\bauthor{\bsnm{Hwang}, \binits{J.D.}},
\bauthor{\bsnm{Bhagavatula}, \binits{C.}},
\bauthor{\bsnm{Bras}, \binits{R.L.}},
\bauthor{\bsnm{Da}, \binits{J.}},
\bauthor{\bsnm{Sakaguchi}, \binits{K.}},
\bauthor{\bsnm{Bosselut}, \binits{A.}},
\bauthor{\bsnm{Choi}, \binits{Y.}}:
\bctitle{(comet-) atomic 2020: On symbolic and neural commonsense knowledge graphs}.
In: \bbtitle{Thirty-Fifth {AAAI} Conference on Artificial Intelligence, {AAAI} 2021, Thirty-Third Conference on Innovative Applications of Artificial Intelligence, {IAAI} 2021, The Eleventh Symposium on Educational Advances in Artificial Intelligence, {EAAI} 2021, Virtual Event, February 2-9, 2021},
pp. \bfpage{6384}--\blpage{6392}
(\byear{2021})
\end{bchapter}
\endbibitem

\bibitem[\protect\citeauthoryear{Jiang et~al.}{2019}]{DBLP:conf/naacl/JiangW019}
\begin{bchapter}
\bauthor{\bsnm{Jiang}, \binits{K.}},
\bauthor{\bsnm{Wu}, \binits{D.}},
\bauthor{\bsnm{Jiang}, \binits{H.}}:
\bctitle{Freebaseqa: {A} new factoid {QA} data set matching trivia-style question-answer pairs with freebase}.
In: \beditor{\bsnm{Burstein}, \binits{J.}},
\beditor{\bsnm{Doran}, \binits{C.}},
\beditor{\bsnm{Solorio}, \binits{T.}} (eds.)
\bbtitle{Proceedings of the 2019 Conference of the North American Chapter of the Association for Computational Linguistics: Human Language Technologies, {NAACL-HLT} 2019, Minneapolis, MN, USA, June 2-7, 2019, Volume 1 (Long and Short Papers)},
pp. \bfpage{318}--\blpage{323}
(\byear{2019}).
\burl{https://doi.org/10.18653/v1/n19-1028}
\end{bchapter}
\endbibitem

\bibitem[\protect\citeauthoryear{Ye et~al.}{2022}]{DBLP:conf/acl/YeL0S22}
\begin{bchapter}
\bauthor{\bsnm{Ye}, \binits{D.}},
\bauthor{\bsnm{Lin}, \binits{Y.}},
\bauthor{\bsnm{Li}, \binits{P.}},
\bauthor{\bsnm{Sun}, \binits{M.}}:
\bctitle{Packed levitated marker for entity and relation extraction}.
In: \beditor{\bsnm{Muresan}, \binits{S.}},
\beditor{\bsnm{Nakov}, \binits{P.}},
\beditor{\bsnm{Villavicencio}, \binits{A.}} (eds.)
\bbtitle{Proceedings of the 60th Annual Meeting of the Association for Computational Linguistics (Volume 1: Long Papers), {ACL} 2022, Dublin, Ireland, May 22-27, 2022},
pp. \bfpage{4904}--\blpage{4917}
(\byear{2022}).
\burl{https://doi.org/10.18653/v1/2022.acl-long.337}
\end{bchapter}
\endbibitem

\bibitem[\protect\citeauthoryear{Park and Kim}{2021}]{DBLP:journals/corr/abs-2107-09332}
\begin{botherref}
\oauthor{\bsnm{Park}, \binits{S.}},
\oauthor{\bsnm{Kim}, \binits{H.}}:
Improving sentence-level relation extraction through curriculum learning.
CoRR
(2021)
\end{botherref}
\endbibitem

\bibitem[\protect\citeauthoryear{Wang et~al.}{2023}]{DBLP:journals/corr/abs-2204-07241}
\begin{bchapter}
\bauthor{\bsnm{Wang}, \binits{S.}},
\bauthor{\bsnm{Yu}, \binits{M.}},
\bauthor{\bsnm{Huang}, \binits{L.}}:
\bctitle{The art of prompting: Event detection based on type specific prompts}.
In: \beditor{\bsnm{Rogers}, \binits{A.}},
\beditor{\bsnm{Boyd{-}Graber}, \binits{J.L.}},
\beditor{\bsnm{Okazaki}, \binits{N.}} (eds.)
\bbtitle{Proceedings of the 61st Annual Meeting of the Association for Computational Linguistics (Volume 2: Short Papers), {ACL} 2023, Toronto, Canada, July 9-14, 2023},
pp. \bfpage{1286}--\blpage{1299}
(\byear{2023}).
\burl{https://doi.org/10.18653/v1/2023.acl-short.111}
\end{bchapter}
\endbibitem

\bibitem[\protect\citeauthoryear{Wang et~al.}{2022}]{wang2022language}
\begin{bchapter}
\bauthor{\bsnm{Wang}, \binits{X.}},
\bauthor{\bsnm{He}, \binits{Q.}},
\bauthor{\bsnm{Liang}, \binits{J.}},
\bauthor{\bsnm{Xiao}, \binits{Y.}}:
\bctitle{Language models as knowledge embeddings}.
In: \beditor{\bsnm{Raedt}, \binits{L.D.}} (ed.)
\bbtitle{Proceedings of the Thirty-First International Joint Conference on Artificial Intelligence, {IJCAI} 2022, Vienna, Austria, 23-29 July 2022},
pp. \bfpage{2291}--\blpage{2297}
(\byear{2022}).
\burl{https://doi.org/10.24963/ijcai.2022/318}
\end{bchapter}
\endbibitem

\bibitem[\protect\citeauthoryear{Hwang et~al.}{2021}]{Hwang2021COMETATOMIC2O}
\begin{bchapter}
\bauthor{\bsnm{Hwang}, \binits{J.D.}},
\bauthor{\bsnm{Bhagavatula}, \binits{C.}},
\bauthor{\bsnm{Bras}, \binits{R.L.}},
\bauthor{\bsnm{Da}, \binits{J.}},
\bauthor{\bsnm{Sakaguchi}, \binits{K.}},
\bauthor{\bsnm{Bosselut}, \binits{A.}},
\bauthor{\bsnm{Choi}, \binits{Y.}}:
\bctitle{(comet-) atomic 2020: On symbolic and neural commonsense knowledge graphs}.
In: \bbtitle{Thirty-Fifth {AAAI} Conference on Artificial Intelligence, {AAAI} 2021, Thirty-Third Conference on Innovative Applications of Artificial Intelligence, {IAAI} 2021, The Eleventh Symposium on Educational Advances in Artificial Intelligence, {EAAI} 2021, Virtual Event, February 2-9, 2021},
pp. \bfpage{6384}--\blpage{6392}
(\byear{2021}).
\burl{https://doi.org/10.1609/aaai.v35i7.16792}
\end{bchapter}
\endbibitem

\bibitem[\protect\citeauthoryear{Yu et~al.}{2023}]{DBLP:journals/corr/abs-2210-00063}
\begin{bchapter}
\bauthor{\bsnm{Yu}, \binits{D.}},
\bauthor{\bsnm{Zhang}, \binits{S.}},
\bauthor{\bsnm{Ng}, \binits{P.}},
\bauthor{\bsnm{Zhu}, \binits{H.}},
\bauthor{\bsnm{Li}, \binits{A.H.}},
\bauthor{\bsnm{Wang}, \binits{J.}},
\bauthor{\bsnm{Hu}, \binits{Y.}},
\bauthor{\bsnm{Wang}, \binits{W.Y.}},
\bauthor{\bsnm{Wang}, \binits{Z.}},
\bauthor{\bsnm{Xiang}, \binits{B.}}:
\bctitle{Decaf: Joint decoding of answers and logical forms for question answering over knowledge bases}.
In: \bbtitle{The Eleventh International Conference on Learning Representations, {ICLR} 2023, Kigali, Rwanda, May 1-5, 2023}
(\byear{2023})
\end{bchapter}
\endbibitem

\bibitem[\protect\citeauthoryear{Madani and Joseph}{2023}]{DBLP:journals/corr/abs-2303-02206}
\begin{bchapter}
\bauthor{\bsnm{Madani}, \binits{N.}},
\bauthor{\bsnm{Joseph}, \binits{K.}}:
\bctitle{Answering questions over knowledge graphs using logic programming along with language models}.
In: \beditor{\bsnm{Maughan}, \binits{K.}},
\beditor{\bsnm{Liu}, \binits{R.}},
\beditor{\bsnm{Burns}, \binits{T.F.}} (eds.)
\bbtitle{The First Tiny Papers Track at {ICLR} 2023, Tiny Papers @ {ICLR} 2023, Kigali, Rwanda, May 5, 2023}
(\byear{2023})
\end{bchapter}
\endbibitem

\bibitem[\protect\citeauthoryear{Gao et~al.}{2023}]{DBLP:journals/corr/abs-2303-03836}
\begin{barticle}
\bauthor{\bsnm{Gao}, \binits{J.}},
\bauthor{\bsnm{Zhao}, \binits{H.}},
\bauthor{\bsnm{Yu}, \binits{C.}},
\bauthor{\bsnm{Xu}, \binits{R.}}:
\batitle{Exploring the feasibility of chatgpt for event extraction}.
\bjtitle{CoRR}
(\byear{2023})
\doiurl{10.48550/arXiv.2303.03836}
\end{barticle}
\endbibitem

\bibitem[\protect\citeauthoryear{Dong et~al.}{2023}]{DBLP:journals/corr/abs-2301-00234}
\begin{barticle}
\bauthor{\bsnm{Dong}, \binits{Q.}},
\bauthor{\bsnm{Li}, \binits{L.}},
\bauthor{\bsnm{Dai}, \binits{D.}},
\bauthor{\bsnm{Zheng}, \binits{C.}},
\bauthor{\bsnm{Wu}, \binits{Z.}},
\bauthor{\bsnm{Chang}, \binits{B.}},
\bauthor{\bsnm{Sun}, \binits{X.}},
\bauthor{\bsnm{Xu}, \binits{J.}},
\bauthor{\bsnm{Li}, \binits{L.}},
\bauthor{\bsnm{Sui}, \binits{Z.}}:
\batitle{A survey for in-context learning}.
\bjtitle{CoRR}
(\byear{2023})
\doiurl{10.48550/ARXIV.2301.00234}
\end{barticle}
\endbibitem

\bibitem[\protect\citeauthoryear{Wei et~al.}{2023}]{DBLP:journals/corr/abs-2303-03846}
\begin{botherref}
\oauthor{\bsnm{Wei}, \binits{J.W.}},
\oauthor{\bsnm{Wei}, \binits{J.}},
\oauthor{\bsnm{Tay}, \binits{Y.}},
\oauthor{\bsnm{Tran}, \binits{D.}},
\oauthor{\bsnm{Webson}, \binits{A.}},
\oauthor{\bsnm{Lu}, \binits{Y.}},
\oauthor{\bsnm{Chen}, \binits{X.}},
\oauthor{\bsnm{Liu}, \binits{H.}},
\oauthor{\bsnm{Huang}, \binits{D.}},
\oauthor{\bsnm{Zhou}, \binits{D.}},
\oauthor{\bsnm{Ma}, \binits{T.}}:
Larger language models do in-context learning differently.
CoRR
(2023)
\end{botherref}
\endbibitem

\bibitem[\protect\citeauthoryear{Wang et~al.}{2024}]{A_Survey_on_LLM_based_Autonomous_Agents}
\begin{barticle}
\bauthor{\bsnm{Wang}, \binits{L.}},
\bauthor{\bsnm{Ma}, \binits{C.}},
\bauthor{\bsnm{Feng}, \binits{X.}},
\bauthor{\bsnm{Zhang}, \binits{Z.}},
\bauthor{\bsnm{Yang}, \binits{H.}},
\bauthor{\bsnm{Zhang}, \binits{J.}},
\bauthor{\bsnm{Chen}, \binits{Z.}},
\bauthor{\bsnm{Tang}, \binits{J.}},
\bauthor{\bsnm{Chen}, \binits{X.}},
\bauthor{\bsnm{Lin}, \binits{Y.}},
\bauthor{\bsnm{Zhao}, \binits{W.X.}},
\bauthor{\bsnm{Wei}, \binits{Z.}},
\bauthor{\bsnm{Wen}, \binits{J.}}:
\batitle{A survey on large language model based autonomous agents}.
\bjtitle{Frontiers Comput. Sci.}
\bvolume{18}(\bissue{6}),
\bfpage{186345}
(\byear{2024})
\doiurl{10.1007/S11704-024-40231-1}
\end{barticle}
\endbibitem

\bibitem[\protect\citeauthoryear{Xi et~al.}{2023}]{The_Rise_and_Potential_of_Agents}
\begin{barticle}
\bauthor{\bsnm{Xi}, \binits{Z.}},
\bauthor{\bsnm{Chen}, \binits{W.}},
\bauthor{\bsnm{Guo}, \binits{X.}},
\bauthor{\bsnm{He}, \binits{W.}},
\bauthor{\bsnm{Ding}, \binits{Y.}},
\bauthor{\bsnm{Hong}, \binits{B.}},
\bauthor{\bsnm{Zhang}, \binits{M.}},
\bauthor{\bsnm{Wang}, \binits{J.}},
\bauthor{\bsnm{Jin}, \binits{S.}},
\bauthor{\bsnm{Zhou}, \binits{E.}},
\bauthor{\bsnm{Zheng}, \binits{R.}},
\bauthor{\bsnm{Fan}, \binits{X.}},
\bauthor{\bsnm{Wang}, \binits{X.}},
\bauthor{\bsnm{Xiong}, \binits{L.}},
\bauthor{\bsnm{Zhou}, \binits{Y.}},
\bauthor{\bsnm{Wang}, \binits{W.}},
\bauthor{\bsnm{Jiang}, \binits{C.}},
\bauthor{\bsnm{Zou}, \binits{Y.}},
\bauthor{\bsnm{Liu}, \binits{X.}},
\bauthor{\bsnm{Yin}, \binits{Z.}},
\bauthor{\bsnm{Dou}, \binits{S.}},
\bauthor{\bsnm{Weng}, \binits{R.}},
\bauthor{\bsnm{Cheng}, \binits{W.}},
\bauthor{\bsnm{Zhang}, \binits{Q.}},
\bauthor{\bsnm{Qin}, \binits{W.}},
\bauthor{\bsnm{Zheng}, \binits{Y.}},
\bauthor{\bsnm{Qiu}, \binits{X.}},
\bauthor{\bsnm{Huan}, \binits{X.}},
\bauthor{\bsnm{Gui}, \binits{T.}}:
\batitle{The rise and potential of large language model based agents: {A} survey}.
\bjtitle{CoRR}
(\byear{2023})
\doiurl{10.48550/arXiv.2309.07864}
\end{barticle}
\endbibitem

\bibitem[\protect\citeauthoryear{Zhao et~al.}{2023}]{An_In_depth_Survey_of_LLM_based_Agents}
\begin{barticle}
\bauthor{\bsnm{Zhao}, \binits{P.}},
\bauthor{\bsnm{Jin}, \binits{Z.}},
\bauthor{\bsnm{Cheng}, \binits{N.}}:
\batitle{An in-depth survey of large language model-based artificial intelligence agents}.
\bjtitle{CoRR}
(\byear{2023})
\doiurl{10.48550/arXiv.2309.14365}
\end{barticle}
\endbibitem

\bibitem[\protect\citeauthoryear{Li et~al.}{2023}]{DBLP:journals/corr/abs-2303-17760}
\begin{bchapter}
\bauthor{\bsnm{Li}, \binits{G.}},
\bauthor{\bsnm{Hammoud}, \binits{H.A.A.K.}},
\bauthor{\bsnm{Itani}, \binits{H.}},
\bauthor{\bsnm{Khizbullin}, \binits{D.}},
\bauthor{\bsnm{Ghanem}, \binits{B.}}:
\bctitle{Camel: Communicative agents for "mind" exploration of large language model society}.
In: \bbtitle{Thirty-seventh Conference on Neural Information Processing Systems}
(\byear{2023})
\end{bchapter}
\endbibitem

\bibitem[\protect\citeauthoryear{Brown et~al.}{2020}]{DBLP:journals/corr/abs-2005-14165}
\begin{bchapter}
\bauthor{\bsnm{Brown}, \binits{T.B.}},
\bauthor{\bsnm{Mann}, \binits{B.}},
\bauthor{\bsnm{Ryder}, \binits{N.}},
\bauthor{\bsnm{Subbiah}, \binits{M.}},
\bauthor{\bsnm{Kaplan}, \binits{J.}},
\bauthor{\bsnm{Dhariwal}, \binits{P.}},
\bauthor{\bsnm{Neelakantan}, \binits{A.}},
\bauthor{\bsnm{Shyam}, \binits{P.}},
\bauthor{\bsnm{Sastry}, \binits{G.}},
\bauthor{\bsnm{Askell}, \binits{A.}},
\bauthor{\bsnm{Agarwal}, \binits{S.}},
\bauthor{\bsnm{Herbert{-}Voss}, \binits{A.}},
\bauthor{\bsnm{Krueger}, \binits{G.}},
\bauthor{\bsnm{Henighan}, \binits{T.}},
\bauthor{\bsnm{Child}, \binits{R.}},
\bauthor{\bsnm{Ramesh}, \binits{A.}},
\bauthor{\bsnm{Ziegler}, \binits{D.M.}},
\bauthor{\bsnm{Wu}, \binits{J.}},
\bauthor{\bsnm{Winter}, \binits{C.}},
\bauthor{\bsnm{Hesse}, \binits{C.}},
\bauthor{\bsnm{Chen}, \binits{M.}},
\bauthor{\bsnm{Sigler}, \binits{E.}},
\bauthor{\bsnm{Litwin}, \binits{M.}},
\bauthor{\bsnm{Gray}, \binits{S.}},
\bauthor{\bsnm{Chess}, \binits{B.}},
\bauthor{\bsnm{Clark}, \binits{J.}},
\bauthor{\bsnm{Berner}, \binits{C.}},
\bauthor{\bsnm{McCandlish}, \binits{S.}},
\bauthor{\bsnm{Radford}, \binits{A.}},
\bauthor{\bsnm{Sutskever}, \binits{I.}},
\bauthor{\bsnm{Amodei}, \binits{D.}}:
\bctitle{Language models are few-shot learners}.
In: \beditor{\bsnm{Larochelle}, \binits{H.}},
\beditor{\bsnm{Ranzato}, \binits{M.}},
\beditor{\bsnm{Hadsell}, \binits{R.}},
\beditor{\bsnm{Balcan}, \binits{M.}},
\beditor{\bsnm{Lin}, \binits{H.}} (eds.)
\bbtitle{Advances in Neural Information Processing Systems 33: Annual Conference on Neural Information Processing Systems 2020, NeurIPS 2020, December 6-12, 2020, Virtual}
(\byear{2020}).
\burl{https://proceedings.neurips.cc/paper/2020/hash/1457c0d6bfcb4967418bfb8ac142f64a-Abstract.html}
\end{bchapter}
\endbibitem

\bibitem[\protect\citeauthoryear{Wei et~al.}{2022}]{DBLP:journals/corr/abs-2206-07682}
\begin{botherref}
\oauthor{\bsnm{Wei}, \binits{J.}},
\oauthor{\bsnm{Tay}, \binits{Y.}},
\oauthor{\bsnm{Bommasani}, \binits{R.}},
\oauthor{\bsnm{Raffel}, \binits{C.}},
\oauthor{\bsnm{Zoph}, \binits{B.}},
\oauthor{\bsnm{Borgeaud}, \binits{S.}},
\oauthor{\bsnm{Yogatama}, \binits{D.}},
\oauthor{\bsnm{Bosma}, \binits{M.}},
\oauthor{\bsnm{Zhou}, \binits{D.}},
\oauthor{\bsnm{Metzler}, \binits{D.}},
\oauthor{\bsnm{Chi}, \binits{E.H.}},
\oauthor{\bsnm{Hashimoto}, \binits{T.}},
\oauthor{\bsnm{Vinyals}, \binits{O.}},
\oauthor{\bsnm{Liang}, \binits{P.}},
\oauthor{\bsnm{Dean}, \binits{J.}},
\oauthor{\bsnm{Fedus}, \binits{W.}}:
Emergent abilities of large language models.
Trans. Mach. Learn. Res.
\textbf{2022}
(2022)
\end{botherref}
\endbibitem

\bibitem[\protect\citeauthoryear{Bang et~al.}{2023}]{DBLP:journals/corr/abs-2302-04023}
\begin{bchapter}
\bauthor{\bsnm{Bang}, \binits{Y.}},
\bauthor{\bsnm{Cahyawijaya}, \binits{S.}},
\bauthor{\bsnm{Lee}, \binits{N.}},
\bauthor{\bsnm{Dai}, \binits{W.}},
\bauthor{\bsnm{Su}, \binits{D.}},
\bauthor{\bsnm{Wilie}, \binits{B.}},
\bauthor{\bsnm{Lovenia}, \binits{H.}},
\bauthor{\bsnm{Ji}, \binits{Z.}},
\bauthor{\bsnm{Yu}, \binits{T.}},
\bauthor{\bsnm{Chung}, \binits{W.}},
\bauthor{\bsnm{Do}, \binits{Q.V.}},
\bauthor{\bsnm{Xu}, \binits{Y.}},
\bauthor{\bsnm{Fung}, \binits{P.}}:
\bctitle{A multitask, multilingual, multimodal evaluation of chatgpt on reasoning, hallucination, and interactivity}.
In: \beditor{\bsnm{Park}, \binits{J.C.}},
\beditor{\bsnm{Arase}, \binits{Y.}},
\beditor{\bsnm{Hu}, \binits{B.}},
\beditor{\bsnm{Lu}, \binits{W.}},
\beditor{\bsnm{Wijaya}, \binits{D.}},
\beditor{\bsnm{Purwarianti}, \binits{A.}},
\beditor{\bsnm{Krisnadhi}, \binits{A.A.}} (eds.)
\bbtitle{Proceedings of the 13th International Joint Conference on Natural Language Processing and the 3rd Conference of the Asia-Pacific Chapter of the Association for Computational Linguistics, {IJCNLP} 2023 -Volume 1: Long Papers, Nusa Dua, Bali, November 1 - 4, 2023},
pp. \bfpage{675}--\blpage{718}
(\byear{2023}).
\burl{https://doi.org/10.18653/v1/2023.ijcnlp-main.45}
\end{bchapter}
\endbibitem

\bibitem[\protect\citeauthoryear{Nori et~al.}{2023}]{nori2023capabilities}
\begin{barticle}
\bauthor{\bsnm{Nori}, \binits{H.}},
\bauthor{\bsnm{King}, \binits{N.}},
\bauthor{\bsnm{McKinney}, \binits{S.M.}},
\bauthor{\bsnm{Carignan}, \binits{D.}},
\bauthor{\bsnm{Horvitz}, \binits{E.}}:
\batitle{Capabilities of {GPT-4} on medical challenge problems}.
\bjtitle{CoRR}
(\byear{2023})
\doiurl{10.48550/ARXIV.2303.13375}
\end{barticle}
\endbibitem

\bibitem[\protect\citeauthoryear{Qiao et~al.}{2023}]{ACL2023_PromptReasoningSurvey}
\begin{bchapter}
\bauthor{\bsnm{Qiao}, \binits{S.}},
\bauthor{\bsnm{Ou}, \binits{Y.}},
\bauthor{\bsnm{Zhang}, \binits{N.}},
\bauthor{\bsnm{Chen}, \binits{X.}},
\bauthor{\bsnm{Yao}, \binits{Y.}},
\bauthor{\bsnm{Deng}, \binits{S.}},
\bauthor{\bsnm{Tan}, \binits{C.}},
\bauthor{\bsnm{Huang}, \binits{F.}},
\bauthor{\bsnm{Chen}, \binits{H.}}:
\bctitle{Reasoning with language model prompting: {A} survey}.
In: \beditor{\bsnm{Rogers}, \binits{A.}},
\beditor{\bsnm{Boyd{-}Graber}, \binits{J.L.}},
\beditor{\bsnm{Okazaki}, \binits{N.}} (eds.)
\bbtitle{Proceedings of the 61st Annual Meeting of the Association for Computational Linguistics (Volume 1: Long Papers), {ACL} 2023, Toronto, Canada, July 9-14, 2023},
pp. \bfpage{5368}--\blpage{5393}
(\byear{2023}).
\burl{https://doi.org/10.18653/v1/2023.acl-long.294}
\end{bchapter}
\endbibitem

\bibitem[\protect\citeauthoryear{S{\'{a}}nchez et~al.}{2023}]{DBLP:conf/acl/SanchezCWCM23}
\begin{bchapter}
\bauthor{\bsnm{S{\'{a}}nchez}, \binits{R.J.}},
\bauthor{\bsnm{Conrads}, \binits{L.}},
\bauthor{\bsnm{Welke}, \binits{P.}},
\bauthor{\bsnm{Cvejoski}, \binits{K.}},
\bauthor{\bsnm{Marin}, \binits{C.O.}}:
\bctitle{Hidden schema networks}.
In: \beditor{\bsnm{Rogers}, \binits{A.}},
\beditor{\bsnm{Boyd{-}Graber}, \binits{J.L.}},
\beditor{\bsnm{Okazaki}, \binits{N.}} (eds.)
\bbtitle{Proceedings of the 61st Annual Meeting of the Association for Computational Linguistics (Volume 1: Long Papers), {ACL} 2023, Toronto, Canada, July 9-14, 2023},
pp. \bfpage{4764}--\blpage{4798}
(\byear{2023}).
\burl{https://doi.org/10.18653/v1/2023.acl-long.263}
\end{bchapter}
\endbibitem

\bibitem[\protect\citeauthoryear{Ma et~al.}{2023}]{DBLP:conf/emnlp/Ma0HS23}
\begin{bchapter}
\bauthor{\bsnm{Ma}, \binits{Y.}},
\bauthor{\bsnm{Cao}, \binits{Y.}},
\bauthor{\bsnm{Hong}, \binits{Y.}},
\bauthor{\bsnm{Sun}, \binits{A.}}:
\bctitle{Large language model is not a good few-shot information extractor, but a good reranker for hard samples!}
In: \beditor{\bsnm{Bouamor}, \binits{H.}},
\beditor{\bsnm{Pino}, \binits{J.}},
\beditor{\bsnm{Bali}, \binits{K.}} (eds.)
\bbtitle{Findings of the Association for Computational Linguistics: {EMNLP} 2023, Singapore, December 6-10, 2023},
pp. \bfpage{10572}--\blpage{10601}
(\byear{2023}).
\burl{https://doi.org/10.18653/v1/2023.findings-emnlp.710}
\end{bchapter}
\endbibitem

\bibitem[\protect\citeauthoryear{Jeblick et~al.}{2022}]{DBLP:journals/corr/abs-2212-14882}
\begin{barticle}
\bauthor{\bsnm{Jeblick}, \binits{K.}},
\bauthor{\bsnm{Schachtner}, \binits{B.}},
\bauthor{\bsnm{Dexl}, \binits{J.}},
\bauthor{\bsnm{Mittermeier}, \binits{A.}},
\bauthor{\bsnm{St{\"{u}}ber}, \binits{A.T.}},
\bauthor{\bsnm{Topalis}, \binits{J.}},
\bauthor{\bsnm{Weber}, \binits{T.}},
\bauthor{\bsnm{Wesp}, \binits{P.}},
\bauthor{\bsnm{Sabel}, \binits{B.O.}},
\bauthor{\bsnm{Ricke}, \binits{J.}},
\bauthor{\bsnm{Ingrisch}, \binits{M.}}:
\batitle{Chatgpt makes medicine easy to swallow: An exploratory case study on simplified radiology reports}.
\bjtitle{CoRR}
(\byear{2022})
\doiurl{10.48550/arXiv.2212.14882}
\end{barticle}
\endbibitem

\bibitem[\protect\citeauthoryear{Tan et~al.}{2023}]{DBLP:journals/corr/abs-2303-07992}
\begin{bchapter}
\bauthor{\bsnm{Tan}, \binits{Y.}},
\bauthor{\bsnm{Min}, \binits{D.}},
\bauthor{\bsnm{Li}, \binits{Y.}},
\bauthor{\bsnm{Li}, \binits{W.}},
\bauthor{\bsnm{Hu}, \binits{N.}},
\bauthor{\bsnm{Chen}, \binits{Y.}},
\bauthor{\bsnm{Qi}, \binits{G.}}:
\bctitle{Can chatgpt replace traditional {KBQA} models? an in-depth analysis of the question answering performance of the {GPT} {LLM} family}.
In: \beditor{\bsnm{Payne}, \binits{T.R.}},
\beditor{\bsnm{Presutti}, \binits{V.}},
\beditor{\bsnm{Qi}, \binits{G.}},
\beditor{\bsnm{Poveda{-}Villal{\'{o}}n}, \binits{M.}},
\beditor{\bsnm{Stoilos}, \binits{G.}},
\beditor{\bsnm{Hollink}, \binits{L.}},
\beditor{\bsnm{Kaoudi}, \binits{Z.}},
\beditor{\bsnm{Cheng}, \binits{G.}},
\beditor{\bsnm{Li}, \binits{J.}} (eds.)
\bbtitle{The Semantic Web - {ISWC} 2023 - 22nd International Semantic Web Conference, Athens, Greece, November 6-10, 2023, Proceedings, Part {I}}.
\bsertitle{Lecture Notes in Computer Science},
vol. \bseriesno{14265},
pp. \bfpage{348}--\blpage{367}
(\byear{2023}).
\burl{https://doi.org/10.1007/978-3-031-47240-4\_19}
\end{bchapter}
\endbibitem

\bibitem[\protect\citeauthoryear{Jiao et~al.}{2023}]{DBLP:journals/corr/abs-2301-08745}
\begin{barticle}
\bauthor{\bsnm{Jiao}, \binits{W.}},
\bauthor{\bsnm{Wang}, \binits{W.}},
\bauthor{\bsnm{Huang}, \binits{J.}},
\bauthor{\bsnm{Wang}, \binits{X.}},
\bauthor{\bsnm{Tu}, \binits{Z.}}:
\batitle{Is chatgpt {A} good translator? {A} preliminary study}.
\bjtitle{CoRR}
(\byear{2023})
\doiurl{10.48550/arXiv.2301.08745}
\end{barticle}
\endbibitem

\bibitem[\protect\citeauthoryear{Kasai et~al.}{2023}]{kasai2023evaluating}
\begin{barticle}
\bauthor{\bsnm{Kasai}, \binits{J.}},
\bauthor{\bsnm{Kasai}, \binits{Y.}},
\bauthor{\bsnm{Sakaguchi}, \binits{K.}},
\bauthor{\bsnm{Yamada}, \binits{Y.}},
\bauthor{\bsnm{Radev}, \binits{D.}}:
\batitle{Evaluating {GPT-4} and chatgpt on japanese medical licensing examinations}.
\bjtitle{CoRR}
(\byear{2023})
\doiurl{10.48550/ARXIV.2303.18027}
\end{barticle}
\endbibitem

\bibitem[\protect\citeauthoryear{Sifatkaur et~al.}{2023}]{DBLP:journals/corr/abs-2303-11436}
\begin{barticle}
\bauthor{\bsnm{Sifatkaur}},
\bauthor{\bsnm{Singh}, \binits{M.}},
\bauthor{\bsnm{B}, \binits{V.S.}},
\bauthor{\bsnm{Malviya}, \binits{N.}}:
\batitle{Mind meets machine: Unravelling gpt-4's cognitive psychology}.
\bjtitle{CoRR}
(\byear{2023})
\doiurl{10.48550/arXiv.2303.11436}
\end{barticle}
\endbibitem

\bibitem[\protect\citeauthoryear{Nunes et~al.}{2023}]{DBLP:journals/corr/abs-2303-17003}
\begin{barticle}
\bauthor{\bsnm{Nunes}, \binits{D.}},
\bauthor{\bsnm{Primi}, \binits{R.}},
\bauthor{\bsnm{Pires}, \binits{R.}},
\bauthor{\bsnm{Alencar~Lotufo}, \binits{R.}},
\bauthor{\bsnm{Nogueira}, \binits{R.F.}}:
\batitle{Evaluating {GPT-3.5} and {GPT-4} models on brazilian university admission exams}.
\bjtitle{CoRR}
(\byear{2023})
\doiurl{10.48550/arXiv.2303.17003}
\end{barticle}
\endbibitem

\bibitem[\protect\citeauthoryear{Lyu et~al.}{2023}]{DBLP:journals/corr/abs-2303-09038}
\begin{barticle}
\bauthor{\bsnm{Lyu}, \binits{Q.}},
\bauthor{\bsnm{Tan}, \binits{J.}},
\bauthor{\bsnm{Zapadka}, \binits{M.E.}},
\bauthor{\bsnm{Ponnatapuram}, \binits{J.}},
\bauthor{\bsnm{Niu}, \binits{C.}},
\bauthor{\bsnm{Wang}, \binits{G.}},
\bauthor{\bsnm{Whitlow}, \binits{C.T.}}:
\batitle{Translating radiology reports into plain language using chatgpt and {GPT-4} with prompt learning: Promising results, limitations, and potential}.
\bjtitle{Vis. Comput. Ind. Biomed. Art 6, 9}
(\byear{2023})
\doiurl{10.1186/s42492-023-00136-5}
\end{barticle}
\endbibitem

\bibitem[\protect\citeauthoryear{Li et~al.}{2024}]{kgc/Contextualization_Distillation}
\begin{bchapter}
\bauthor{\bsnm{Li}, \binits{D.}},
\bauthor{\bsnm{Tan}, \binits{Z.}},
\bauthor{\bsnm{Chen}, \binits{T.}},
\bauthor{\bsnm{Liu}, \binits{H.}}:
\bctitle{Contextualization distillation from large language model for knowledge graph completion}.
In: \beditor{\bsnm{Graham}, \binits{Y.}},
\beditor{\bsnm{Purver}, \binits{M.}} (eds.)
\bbtitle{Findings of the Association for Computational Linguistics: {EACL} 2024, St. Julian's, Malta, March 17-22, 2024},
pp. \bfpage{458}--\blpage{477}
(\byear{2024}).
\burl{https://aclanthology.org/2024.findings-eacl.32}
\end{bchapter}
\endbibitem

\bibitem[\protect\citeauthoryear{Li et~al.}{2021}]{ner/span-bert}
\begin{bchapter}
\bauthor{\bsnm{Li}, \binits{F.}},
\bauthor{\bsnm{Lin}, \binits{Z.}},
\bauthor{\bsnm{Zhang}, \binits{M.}},
\bauthor{\bsnm{Ji}, \binits{D.}}:
\bctitle{A span-based model for joint overlapped and discontinuous named entity recognition}.
In: \beditor{\bsnm{Zong}, \binits{C.}},
\beditor{\bsnm{Xia}, \binits{F.}},
\beditor{\bsnm{Li}, \binits{W.}},
\beditor{\bsnm{Navigli}, \binits{R.}} (eds.)
\bbtitle{Proceedings of the 59th Annual Meeting of the Association for Computational Linguistics and the 11th International Joint Conference on Natural Language Processing, {ACL/IJCNLP} 2021, (Volume 1: Long Papers), Virtual Event, August 1-6, 2021},
pp. \bfpage{4814}--\blpage{4828}
(\byear{2021}).
\burl{https://doi.org/10.18653/v1/2021.acl-long.372}
\end{bchapter}
\endbibitem

\bibitem[\protect\citeauthoryear{Zhou et~al.}{2024}]{ner/UniversalNER}
\begin{bchapter}
\bauthor{\bsnm{Zhou}, \binits{W.}},
\bauthor{\bsnm{Zhang}, \binits{S.}},
\bauthor{\bsnm{Gu}, \binits{Y.}},
\bauthor{\bsnm{Chen}, \binits{M.}},
\bauthor{\bsnm{Poon}, \binits{H.}}:
\bctitle{Universalner: Targeted distillation from large language models for open named entity recognition}.
In: \bbtitle{The Twelfth International Conference on Learning Representations, {ICLR} 2024}
(\byear{2024}).
\burl{https://openreview.net/forum?id=r65xfUb76p}
\end{bchapter}
\endbibitem

\bibitem[\protect\citeauthoryear{Jiang et~al.}{2024}]{re/genres}
\begin{bchapter}
\bauthor{\bsnm{Jiang}, \binits{P.}},
\bauthor{\bsnm{Lin}, \binits{J.}},
\bauthor{\bsnm{Wang}, \binits{Z.}},
\bauthor{\bsnm{Sun}, \binits{J.}},
\bauthor{\bsnm{Han}, \binits{J.}}:
\bctitle{{G}en{RES}: Rethinking evaluation for generative relation extraction in the era of large language models}.
In: \beditor{\bsnm{Duh}, \binits{K.}},
\beditor{\bsnm{Gomez}, \binits{H.}},
\beditor{\bsnm{Bethard}, \binits{S.}} (eds.)
\bbtitle{Proceedings of the 2024 Conference of the North American Chapter of the Association for Computational Linguistics: Human Language Technologies (Volume 1: Long Papers)},
pp. \bfpage{2820}--\blpage{2837}.
\bpublisher{Association for Computational Linguistics},
\blocation{Mexico City, Mexico}
(\byear{2024}).
\burl{https://aclanthology.org/2024.naacl-long.155}
\end{bchapter}
\endbibitem

\bibitem[\protect\citeauthoryear{Wang et~al.}{2022}]{lp/simkgc}
\begin{bchapter}
\bauthor{\bsnm{Wang}, \binits{L.}},
\bauthor{\bsnm{Zhao}, \binits{W.}},
\bauthor{\bsnm{Wei}, \binits{Z.}},
\bauthor{\bsnm{Liu}, \binits{J.}}:
\bctitle{Simkgc: Simple contrastive knowledge graph completion with pre-trained language models}.
In: \beditor{\bsnm{Muresan}, \binits{S.}},
\beditor{\bsnm{Nakov}, \binits{P.}},
\beditor{\bsnm{Villavicencio}, \binits{A.}} (eds.)
\bbtitle{Proceedings of the 60th Annual Meeting of the Association for Computational Linguistics (Volume 1: Long Papers), {ACL} 2022, Dublin, Ireland, May 22-27, 2022},
pp. \bfpage{4281}--\blpage{4294}
(\byear{2022}).
\burl{https://doi.org/10.18653/v1/2022.acl-long.295}
\end{bchapter}
\endbibitem

\bibitem[\protect\citeauthoryear{Li et~al.}{2023}]{lp/lpbert}
\begin{barticle}
\bauthor{\bsnm{Li}, \binits{D.}},
\bauthor{\bsnm{Zhu}, \binits{B.}},
\bauthor{\bsnm{Yang}, \binits{S.}},
\bauthor{\bsnm{Xu}, \binits{K.}},
\bauthor{\bsnm{Yi}, \binits{M.}},
\bauthor{\bsnm{He}, \binits{Y.}},
\bauthor{\bsnm{Wang}, \binits{H.}}:
\batitle{Multi-task pre-training language model for semantic network completion}.
\bjtitle{{ACM} Trans. Asian Low Resour. Lang. Inf. Process.}
\bvolume{22}(\bissue{11}),
\bfpage{250}--\blpage{125020}
(\byear{2023})
\doiurl{10.1145/3627704}
\end{barticle}
\endbibitem

\bibitem[\protect\citeauthoryear{Shu et~al.}{2024}]{lp/kgllm}
\begin{barticle}
\bauthor{\bsnm{Shu}, \binits{D.}},
\bauthor{\bsnm{Chen}, \binits{T.}},
\bauthor{\bsnm{Jin}, \binits{M.}},
\bauthor{\bsnm{Zhang}, \binits{Y.}},
\bauthor{\bsnm{Zhang}, \binits{C.}},
\bauthor{\bsnm{Du}, \binits{M.}},
\bauthor{\bsnm{Zhang}, \binits{Y.}}:
\batitle{Knowledge graph large language model {(KG-LLM)} for link prediction}.
\bjtitle{CoRR}
(\byear{2024})
\doiurl{10.48550/ARXIV.2403.07311}
\end{barticle}
\endbibitem

\bibitem[\protect\citeauthoryear{Hao et~al.}{2023}]{bertnet}
\begin{bchapter}
\bauthor{\bsnm{Hao}, \binits{S.}},
\bauthor{\bsnm{Tan}, \binits{B.}},
\bauthor{\bsnm{Tang}, \binits{K.}},
\bauthor{\bsnm{Ni}, \binits{B.}},
\bauthor{\bsnm{Shao}, \binits{X.}},
\bauthor{\bsnm{Zhang}, \binits{H.}},
\bauthor{\bsnm{Xing}, \binits{E.P.}},
\bauthor{\bsnm{Hu}, \binits{Z.}}:
\bctitle{Bertnet: Harvesting knowledge graphs with arbitrary relations from pretrained language models}.
In: \beditor{\bsnm{Rogers}, \binits{A.}},
\beditor{\bsnm{Boyd{-}Graber}, \binits{J.L.}},
\beditor{\bsnm{Okazaki}, \binits{N.}} (eds.)
\bbtitle{Findings of the Association for Computational Linguistics: {ACL} 2023, Toronto, Canada, July 9-14, 2023},
pp. \bfpage{5000}--\blpage{5015}
(\byear{2023}).
\burl{https://doi.org/10.18653/v1/2023.findings-acl.309}
\end{bchapter}
\endbibitem

\bibitem[\protect\citeauthoryear{Petroni et~al.}{2019}]{LLMasKB}
\begin{bchapter}
\bauthor{\bsnm{Petroni}, \binits{F.}},
\bauthor{\bsnm{Rockt{\"{a}}schel}, \binits{T.}},
\bauthor{\bsnm{Riedel}, \binits{S.}},
\bauthor{\bsnm{Lewis}, \binits{P.S.H.}},
\bauthor{\bsnm{Bakhtin}, \binits{A.}},
\bauthor{\bsnm{Wu}, \binits{Y.}},
\bauthor{\bsnm{Miller}, \binits{A.H.}}:
\bctitle{Language models as knowledge bases?}
In: \beditor{\bsnm{Inui}, \binits{K.}},
\beditor{\bsnm{Jiang}, \binits{J.}},
\beditor{\bsnm{Ng}, \binits{V.}},
\beditor{\bsnm{Wan}, \binits{X.}} (eds.)
\bbtitle{Proceedings of the 2019 Conference on Empirical Methods in Natural Language Processing and the 9th International Joint Conference on Natural Language Processing, {EMNLP-IJCNLP} 2019, Hong Kong, China, November 3-7, 2019},
pp. \bfpage{2463}--\blpage{2473}
(\byear{2019}).
\burl{https://doi.org/10.18653/v1/D19-1250}
\end{bchapter}
\endbibitem

\bibitem[\protect\citeauthoryear{AlKhamissi et~al.}{2022}]{surveyonllmaskb}
\begin{barticle}
\bauthor{\bsnm{AlKhamissi}, \binits{B.}},
\bauthor{\bsnm{Li}, \binits{M.}},
\bauthor{\bsnm{Celikyilmaz}, \binits{A.}},
\bauthor{\bsnm{Diab}, \binits{M.T.}},
\bauthor{\bsnm{Ghazvininejad}, \binits{M.}}:
\batitle{A review on language models as knowledge bases}.
\bjtitle{CoRR}
(\byear{2022})
\doiurl{10.48550/ARXIV.2204.06031}
\end{barticle}
\endbibitem

\bibitem[\protect\citeauthoryear{West et~al.}{2022}]{kgc/symbolicdistill}
\begin{bchapter}
\bauthor{\bsnm{West}, \binits{P.}},
\bauthor{\bsnm{Bhagavatula}, \binits{C.}},
\bauthor{\bsnm{Hessel}, \binits{J.}},
\bauthor{\bsnm{Hwang}, \binits{J.D.}},
\bauthor{\bsnm{Jiang}, \binits{L.}},
\bauthor{\bsnm{Bras}, \binits{R.L.}},
\bauthor{\bsnm{Lu}, \binits{X.}},
\bauthor{\bsnm{Welleck}, \binits{S.}},
\bauthor{\bsnm{Choi}, \binits{Y.}}:
\bctitle{Symbolic knowledge distillation: from general language models to commonsense models}.
In: \beditor{\bsnm{Carpuat}, \binits{M.}},
\beditor{\bsnm{Marneffe}, \binits{M.}},
\beditor{\bsnm{Ru{\'{\i}}z}, \binits{I.V.M.}} (eds.)
\bbtitle{Proceedings of the 2022 Conference of the North American Chapter of the Association for Computational Linguistics: Human Language Technologies, {NAACL} 2022, Seattle, WA, United States, July 10-15, 2022},
pp. \bfpage{4602}--\blpage{4625}
(\byear{2022}).
\burl{https://doi.org/10.18653/v1/2022.naacl-main.341}
\end{bchapter}
\endbibitem

\bibitem[\protect\citeauthoryear{Luo et~al.}{2023}]{kgc/chatrule}
\begin{barticle}
\bauthor{\bsnm{Luo}, \binits{L.}},
\bauthor{\bsnm{Ju}, \binits{J.}},
\bauthor{\bsnm{Xiong}, \binits{B.}},
\bauthor{\bsnm{Li}, \binits{Y.}},
\bauthor{\bsnm{Haffari}, \binits{G.}},
\bauthor{\bsnm{Pan}, \binits{S.}}:
\batitle{Chatrule: Mining logical rules with large language models for knowledge graph reasoning}.
\bjtitle{CoRR}
(\byear{2023})
\doiurl{10.48550/ARXIV.2309.01538}
\end{barticle}
\endbibitem

\bibitem[\protect\citeauthoryear{Miller et~al.}{2016}]{DBLP:conf/emnlp/MillerFDKBW16}
\begin{bchapter}
\bauthor{\bsnm{Miller}, \binits{A.H.}},
\bauthor{\bsnm{Fisch}, \binits{A.}},
\bauthor{\bsnm{Dodge}, \binits{J.}},
\bauthor{\bsnm{Karimi}, \binits{A.}},
\bauthor{\bsnm{Bordes}, \binits{A.}},
\bauthor{\bsnm{Weston}, \binits{J.}}:
\bctitle{Key-value memory networks for directly reading documents}.
In: \beditor{\bsnm{Su}, \binits{J.}},
\beditor{\bsnm{Carreras}, \binits{X.}},
\beditor{\bsnm{Duh}, \binits{K.}} (eds.)
\bbtitle{Proceedings of the 2016 Conference on Empirical Methods in Natural Language Processing, {EMNLP} 2016, Austin, Texas, USA, November 1-4, 2016},
pp. \bfpage{1400}--\blpage{1409}
(\byear{2016}).
\burl{https://doi.org/10.18653/v1/d16-1147}
\end{bchapter}
\endbibitem

\end{thebibliography}

\begin{appendices}

\section{Related Work}

\subsection{Large Language Models}
LLMs are pre-trained on substantial amounts of textual data and have become a significant component of contemporary NLP research.
Recent advancements in NLP have led to the development of highly capable LLMs, such as GPT-3~\citep{DBLP:journals/corr/abs-2005-14165}, ChatGPT, and GPT-4, which exhibit exceptional performance across a diverse array of NLP tasks, including machine translation, text summarization, and question answering.
Concurrently, several previous studies have indicated that LLMs can achieve remarkable results in relevant downstream tasks with minimal or even no demonstration in the prompt~\citep{DBLP:journals/corr/abs-2206-07682, DBLP:journals/corr/abs-2302-04023, nori2023capabilities,ACL2023_PromptReasoningSurvey,DBLP:journals/corr/abs-2302-10205}.
\citet{DBLP:conf/acl/SanchezCWCM23} proposes a novel neural language model that incorporates inductive biases to enforce explicit relational structures. 
of pretrained language models.
 This provides further evidence of the robustness and generality of LLMs.

\subsection{ChatGPT \& GPT-4}
ChatGPT, an advanced LLM developed by OpenAI, is primarily designed for engaging in human-like conversations.
During the fine-tuning process, ChatGPT utilizes RLHF~\citep{DBLP:conf/nips/ChristianoLBMLA17}, thereby enhancing its alignment with human preferences and values.

As a cutting-edge large language model developed by OpenAI, GPT-4 is building upon the successes of its predecessors like GPT-3 and ChatGPT.
Trained on an unparalleled scale of computation and data, it exhibits remarkable generalization, inference, and problem-solving capabilities across diverse domains.
In addition, as a large-scale multimodal model, GPT-4 is capable of processing both image and text inputs.
In general, the public release of GPT-4 offers fresh insights into the future advancement of LLMs and presents novel opportunities and challenges within the realm of NLP.

With the popularity of LLMs, an increasing number of researchers are exploring the specific emergent capabilities and advantages they possess~\citep{DBLP:conf/emnlp/Ma0HS23}.
\citet{DBLP:journals/corr/abs-2302-04023} performs the in-depth analysis of ChatGPT on the multitask, multilingual and multimodal aspects.
The findings indicate that ChatGPT excels at zero-shot learning across various tasks, even outperforming fine-tuned models in certain cases.
However, it faces challenges when generalized to low-resource languages.
Furthermore, in terms of multi-modality, compared to more advanced vision-language models, the capabilities of ChatGPT remain fundamental.
Moreover, ChatGPT has garnered considerable attention in other various domains, including information extraction~\citep{DBLP:journals/corr/abs-2303-03836,DBLP:journals/corr/abs-2302-10205}, reasoning~\citep{reason/nlptasksolver}, text summarization~\citep{DBLP:journals/corr/abs-2212-14882}, question answering \citep{DBLP:journals/corr/abs-2303-07992} and machine translation~\citep{DBLP:journals/corr/abs-2301-08745}, showcasing its versatility and applicability in the broader field of natural language processing.

While there is a growing body of research on ChatGPT, investigations into GPT-4 continue to be relatively limited.
\citet{nori2023capabilities} conducts an extensive assessment of GPT-4 on medical competency examinations and benchmark datasets and shows that GPT-4, without any specialized prompt crafting, surpasses the passing score by over 20 points.
\citet{kasai2023evaluating} also studies GPT-4's performance on the Japanese national medical licensing examinations. 
Furthermore, there are some studies on GPT-4 that focus on cognitive psychology~\citep{DBLP:journals/corr/abs-2303-11436}, academic exams~\citep{DBLP:journals/corr/abs-2303-17003}, and translation of radiology reports~\citep{DBLP:journals/corr/abs-2303-09038}.

\subsection{LLMs for KG}
Now many studies leverage large language models to facilitate the construction of knowledge graphs\citep{kgc/Contextualization_Distillation}.
Some of these tasks focus on specific subtasks within KG construction. 
For instance, LLMs are utilized for named entity recognition and classification~\citep{ner/span-bert,ner/UniversalNER}, leveraging their contextual understanding and linguistic knowledge.
Furthermore, LLMs have also demonstrated utility in tasks such as relation extraction~\citep{DBLP:journals/corr/abs-2302-10205, re/genres} and link prediction~\citep{lp/simkgc,lp/lpbert, lp/kgllm}.
In line with our approach, several studies have explored the use of LLMs as knowledge bases~\citep{bertnet,LLMasKB,surveyonllmaskb,kgc/Contextualization_Distillation} to support KG construction.
For example, some researchers~\citep{kgc/symbolicdistill} propose a symbolic knowledge distillation framework that extracts symbolic knowledge from LLMs.
They first extract commonsense facts from large LLMs like GPT-3, fine-tune smaller student LLMs, and then use these student models to generate KGs. 
Concurrently, ChatRule~\citep{kgc/chatrule} uses LLMs to mine logical rules from KGs, addressing computational intensity and scalability issues present in existing methods. 
ChatRule generates rules with LLMs, integrating the semantic and structural information of KGs, and employs a rule ranking module to evaluate rule quality. 
These studies highlight the extensive potential of LLMs in KG construction, promoting the automation and intelligent development of this field.

\section{Datasets}
\label{sec:datasets}
\textit{\textbf{Entity, Relation and Event Extraction.}}
DuIE2.0~\citep{DBLP:conf/nlpcc/LiHSJLJZLZ19} is a substantial Chinese relationship extraction dataset with more than 210,000 sentences and 48 predefined relationship categories. 
SciERC~\citep{DBLP:conf/emnlp/LuanHOH18} is a collection of scientific abstracts annotated with seven relations. 
Re-TACRED~\citep{DBLP:conf/aaai/StoicaPP21}, an upgraded version of the TACRED dataset, includes over 91,000 sentences across 40 relations. 
MAVEN~\citep{DBLP:conf/emnlp/WangWHJHLLLLZ20} a general-domain event extraction benchmark containing 4,480 documents and 168 event types.

\textit{\textbf{Link Prediction.}}
FB15K-237~\citep{DBLP:conf/emnlp/ToutanovaCPPCG15} is widely used as a benchmark for assessing the performance of knowledge graph embedding model on link prediction, encompassing 237 relations and 14,541 entities.
ATOMIC 2020~\citep{DBLP:conf/aaai/HwangBBDSBC21} serves as a comprehensive commonsense repository with 1.33 million inferential knowledge tuples about entities and events.

\textit{\textbf{Question Answering.}}
FreebaseQA~\citep{DBLP:conf/naacl/JiangW019} is an open-domain QA dataset built on the Freebase knowledge graph, comprising various sourced question-answer pairs.
MetaQA~\citep{DBLP:conf/aaai/ZhangDKSS18}, expanded from WikiMovies~\citep{DBLP:conf/emnlp/MillerFDKBW16}, provides a substantial collection of single-hop and multi-hop question-answer pairs, surpassing 400,000 in total.

\section{Data Collection of VINE}
\label{sec:data_collect}
Using GPT-4 data up to September 2021 as a basis, we select a portion of participants' responses from two competitions organized by \textit{the New York Times} as part of our data sources. 
These competitions include the "February Vocabulary Challenge: Invent a Word"\footnote{\url{https://www.nytimes.com/2022/01/31/learning/february-vocabulary-challenge-invent-a-word.html}}  held in January 2022 and the "Student Vocabulary Challenge: Invent a Word"\footnote{\url{https://www.nytimes.com/2023/02/01/learning/student-vocabulary-challenge-invent-a-word.html}} conducted in February 2023. 
Both competitions aim to promote the creation of distinctive and memorable new words that address gaps in the English language.

Our constructed dataset includes 1,400 sentences, 39 novel relations, and 786 unique entities. 
In the construction process, we ensure that each relation type had a minimum of 10 associated samples to facilitate subsequent experiments. 
Notably, we find that in the Re-TACRED test set, certain types of relations have fewer than 10 corresponding data instances. 
To better conduct our experiments, we select sentences of corresponding types from the training set to offset this deficiency.

\section{Prompts for Evaluation}
\label{sec:Prompts}
Here we list the prompts used in each task during the experiment.
\begin{table*}[!htbp]
 \centering
  \caption{Examples of zero-shot and one-shot prompts we used on Relation Extraction}
 \scalebox{0.75}{
 \begin{tabular}{p{3cm}|p{7cm}|p{7cm}}\toprule
   \textbf{Tasks}   & \textbf{Zero-shot Prompt} & \textbf{One-shot Prompt}\\  \midrule
    Relation Extraction\newline(SciERC)  &
    The list of predicates: ['HYPONYM-OF', 'USED-FOR', 'PART-OF', 'FEATURE-OF', 'COMPARE', 'CONJUNCTION', 'EVALUATE-FOR'].\newline
What Subject-Predicate-Object triples are included in the following sentence? Please return the possible answers according to the list above. Require the answer only in the form : [subject, predicate, object].\newline
The given sentence is: On the internal side, liaisons are established between elements of the text and the graph by using broadly available resources such as a LO-English or better a L0-UNL dictionary, a morphosyntactic parser of L0, and a canonical graph2tree transformation.\newline
Triples:    
&
The list of predicates: ['HYPONYM-OF', 'USED-FOR', 'PART-OF', 'FEATURE-OF', 'COMPARE', 'CONJUNCTION', 'EVALUATE-FOR'].\newline
What Subject-Predicate-Object triples are included in the following sentence? Please return the possible answers according to the list above. Require the answer only in the form: [subject, predicate, object]\newline\newline
Example: \newline
The given sentence is :  We show that various features based on the structure of email-threads can be used to improve upon lexical similarity of discourse segments for question-answer pairing . \newline
Triples: [lexical similarity , FEATURE-OF, discourse segments]\newline\newline
The given sentence is : On the internal side, liaisons are established between elements of the text and the graph by using broadly available resources such as a LO-English or better a L0-UNL dictionary, a morphosyntactic parser of L0, and a canonical graph2tree transformation .\newline
Triples: 
\\   \midrule

  Relation Extraction\newline(DuIE2.0) &\begin{CJK}{UTF8}{gbsn} 已知候选谓词列表： ['董事长', '获奖', '饰演', '成立日期', '母亲', '作者', '歌手', '注册资本', '面积', '父亲', '首都', '人口数量', '代言人', '朝代', '所属专辑', '邮政编码', '主演', '上映时间', '丈夫', '祖籍', '国籍', '简称', '海拔', '出品公司', '主持人', '作曲', '编剧', '妻子', '毕业院校', '总部地点', '所在城市', '校长', '主角', '票房', '主题曲', '制片人', '嘉宾', '作词', '号', '配音', '占地面积', '创始人', '改编自', '气候', '导演', '官方语言', '专业代码', '修业年限'] .\newline
请从以下文本中提取可能的主语-谓语-宾语三元组(SPO三元组)，并以[[主语，谓语，宾语]，...]的形式回答\newline
给定句子： 史奎英，女，中石油下属单位基层退休干部，原国资委主任、中石油董事长蒋洁敏妻子.\newline
SPO三元组:  \end{CJK}& \begin{CJK}{UTF8}{gbsn}已知候选谓词列表： ['主演', '配音', '成立日期', '毕业院校', '父亲', '出品公司', '作词', '作曲', '国籍', '票房', '代言人', '董事长', '朝代', '主持人', '嘉宾', '改编自', '面积', '丈夫', '祖籍', '作者', '号', '主题曲', '专业代码', '主角', '妻子', '导演', '注册资本', '邮政编码', '上映时间', '所属专辑', '获奖', '气候', '简称', '占地面积', '总部地点', '编剧', '所在城市', '首都', '海拔', '官方语言', '校长', '饰演', '修业年限', '人口数量', '创始人', '制片人', '歌手', '母亲'] .\newline
请从以下文本中提取可能的主语-谓语-宾语三元组(SPO三元组)，并以[[主语，谓语，宾语]，...]的形式回答\newline\newline
例如: \newline
给定句子: 641年3月2日文成公主入藏，与松赞干布和亲. \newline
SPO三元组: [松赞干布 , 妻子 , 文成公主 ]、[文成公主 , 丈夫 , 松赞干布 ]\newline\newline
给定句子: 史奎英，女，中石油下属单位基层退休干部，原国资委主任、中石油董事长蒋洁敏妻子\newline
SPO三元组: 
\end{CJK}\\   \midrule
 \end{tabular}
 }
 \label{tab:KG-prompts-RE}
\end{table*}

\begin{table*}[!htbp]
 \centering
  \caption{Examples of zero-shot and one-shot prompts we used on Event Detection, Link Prediction, and Question Answering}
 \scalebox{0.85}{
 \begin{tabular}{p{3cm}|p{6cm}|p{6cm}}\toprule
   \textbf{Tasks}   & \textbf{Zero-shot Prompt} & \textbf{One-shot Prompt}\\  \midrule
    Event Detection  &The list of event types: [......]\newline
Give a sentence: Both teams progressed to the knockout stages by finishing top of their group.\newline
What types of events are included in this sentence? Please return the most likely answer according to the list of event types above. Require the answer in the form: Event type\newline
Ans:\newline
&
The list of event types: [......]\newline
What types of events are included in the following sentence? Please return the most likely answer according to the list of event types above. Require the answer in the form: Event type\newline
Example: \newline
Give a sentence: Unprepared for the attack, the Swedish attempted to save their ships by cutting their anchor ropes and to flee.\newline
Event type: Removing, Rescuing, Escaping, Attack, Self$\_$motion \newline\newline
Give a sentence: Both teams progressed to the knockout stages by finishing top of their group.\newline
Event type: \newline
\\   \midrule
    Link Prediction  &

predict the tail entity [MASK] from the given (40th Academy Awards, time event locations, [MASK]) by completing the sentence "what is the locations of 40th Academy Awards? The answer is".
&
    
predict the tail entity [MASK] from the given (1992 NCAA Men's Division I Basketball Tournament, time event locations, [MASK]) by completing the sentence "what is the locations of 1992 NCAA Men's Division I Basketball Tournament? The answer is ".The answer is Albuquerque, so the [MASK] is Albuquerque. \newline
predict the tail entity [MASK] from the given (40th Academy Awards, time event locations, [MASK]) by completing the sentence "what is the locations of 40th Academy Awards? The answer is".
\\   \midrule
    Question Answering  &
Please answer the following question. Note that there may be more than one answer to the question. 
\newline
Question: [Lamont Johnson] was the director of which films ?
\newline
Answer:
&
Please answer the following question. Note that there may be more than one answer to the question.
\newline
Question: [Aaron Lipstadt] was the director of which movies ?
\newline
Answer: Android | City Limits
\newline
Question: [Lamont Johnson] was the director of which films ?
\newline
Answer:
\\   \midrule
 \end{tabular}
 }
 \label{tab:KG-prompts-ED、LP、QA}
\end{table*}

\onecolumn
\section{Prompts for Virtual Knowledge Extraction}
\begin{table*}[!htbp]
 \centering
 \caption{Examples of Virtual Knowledge Extraction}
 \scalebox{1}{
 \begin{tabular}{p{12cm}}\toprule
   \textbf{Prompts}\\  \midrule
   There might be Subject-Predicate-Object triples in the following sentence. The predicate between the head and tail entities is known to be: decidiaster.\newline
Please find these two entities and give your answers in the form of [subject, predicate, object].\newline\newline
Example: \newline
The given sentence is : The link to the blog was posted on the website of the newspaper Brabants Dagblad , which has identified the crash survivor as Schoolnogo , who had been on safari with his 40-year-old father Patrick , mother Trudy , 41 , and brother Reptance , 11 .\newline
Triples: [Schoolnogo, decidiaster, Reptance]\newline
The given sentence is : Intranguish 's brother , Nugculous , told reporters in Italy that `` there were moments that I believed he would never come back , '' ANSA reported .\newline
Triples: [Intranguish, decidiaster, Nugculous]\newline\newline
The given sentence is : The Dutch newspaper Brabants Dagblad said Adrenaddict had been on safari in South Africa with his mother Trudy , 41 , father Patrick , 40 , and brother Reptance .\newline
Triples: \\\bottomrule
 \end{tabular}
 }
 \label{tab:KG-prompts-VE}
\end{table*}

\end{appendices}


\end{document}